\documentclass[10pt,twocolumn,letterpaper]{article}

\usepackage{cvpr}              % To produce the CAMERA-READY version
\usepackage[table]{xcolor}
\usepackage{multirow}  
\usepackage{graphicx}   % \resizebox, \rotatebox
\usepackage{booktabs}   % \toprule 等
\usepackage{makecell}   % \makecell
\usepackage{booktabs}
\usepackage{multirow}
\usepackage{subcaption}
\usepackage{float}
\newcommand{\spademark}{\ensuremath{\spadesuit}}
\usepackage{pifont}\newcommand{\cmark}{\ding{51}}\newcommand{\xmark}{\ding{55}}
\newcommand{\usym}[1]{%
  \ifnum#1=2713 \cmark\else
  \ifnum#1=2717 \xmark\else #1\fi\fi}
\definecolor{cvprblue}{rgb}{0.21,0.49,0.74}
\definecolor{navyblue}{HTML}{0071BC}
\definecolor{hotpink}{HTML}{FF0080}

\definecolor{oai-white}{HTML}{FFFFFF}
\definecolor{oai-black}{HTML}{000000}
\definecolor{oai-red}{HTML}{FF4500}
\definecolor{oai-green}{HTML}{51DA4C}
\definecolor{oai-blue}{HTML}{0000FF}
\definecolor{oai-yellow}{HTML}{FFF639}
\definecolor{oai-magenta}{HTML}{FF45FF}
\definecolor{oai-cyan}{HTML}{00FFFF}
\definecolor{oai-orange}{HTML}{FE7600}
\definecolor{oai-violet}{HTML}{8A2BE2}
\definecolor{oai-brown}{HTML}{A0522D}
\definecolor{oai-green-050}{HTML}{F4FFF4}
\definecolor{oai-green-100}{HTML}{E9FFE8}
\definecolor{oai-green-200}{HTML}{D9FFD8}
\definecolor{oai-green-300}{HTML}{C9FFC7}
\definecolor{oai-green-400}{HTML}{A6FFA3}
\definecolor{oai-green-500}{HTML}{7CF178}
\definecolor{oai-green-600}{HTML}{51DA4C}
\definecolor{oai-green-700}{HTML}{3FA93B}
\definecolor{oai-green-800}{HTML}{2D712A}
\definecolor{oai-green-900}{HTML}{193718}
\definecolor{oai-gray-000}{HTML}{FFFFFF}
\definecolor{oai-gray-100}{HTML}{FAFAFA}
\definecolor{oai-gray-200}{HTML}{F5F5F5}
\definecolor{oai-gray-300}{HTML}{E5E5E5}
\definecolor{oai-gray-400}{HTML}{FFB7A4}
\definecolor{oai-gray-500}{HTML}{CDCDCD}
\definecolor{oai-gray-600}{HTML}{A8A8A8}
\definecolor{oai-gray-700}{HTML}{747474}
\definecolor{oai-gray-800}{HTML}{393939}
\definecolor{oai-gray-900}{HTML}{000000}
\definecolor{visual}{HTML}{A50E0E}       
\definecolor{linguistic}{HTML}{174EA6}   
\definecolor{relational}{HTML}{E37400}   
\definecolor{egocentric}{HTML}{0D652D}  
\usepackage[pagebackref,breaklinks,colorlinks,allcolors=cvprblue]{hyperref}
\usepackage{mathrsfs}
\usepackage{placeins}
\usepackage{stfloats}
\FloatBarrier
\definecolor{lightbluishgray}{HTML}{291fb2}
\definecolor{mypink}{rgb}{1,0.4,0.7} % Example pink
\def\paperID{7412} % *** Enter the Paper ID here
\def\confName{CVPR}
\def\confYear{2026}
\newcolumntype{Y}{>{\centering\arraybackslash}X}
\usepackage{tabularx}
\usepackage{array}        % 为 \newcolumntype 和 \arraybackslash
\usepackage{booktabs}     % \cmidrule
\usepackage{makecell}     % \Xhline, \makecell
\usepackage{multirow}
\usepackage[table]{xcolor}
\usepackage{colortbl}     % \rowcolor, \cellcolor

\newcolumntype{Y}{>{\centering\arraybackslash}X}

\definecolor{orange}{RGB}{255,165,0}
\colorlet{light-orange}{orange!35}
\definecolor{lightblue}{RGB}{210,230,255}
\definecolor{orange}{RGB}{255,165,0}
\colorlet{light-orange}{orange!35}
\definecolor{lightblue}{RGB}{210,230,255}
\title{Thinking in Dynamics: How Multimodal Large Language Models \\ 
Perceive, Track, and Reason Dynamics in Physical 4D World}

\author{
  Yuzhi Huang\textsuperscript{*\spademark\,1}\quad
  Kairun Wen\textsuperscript{*\,1}\quad
  Rongxin Gao\textsuperscript{*\,1}\quad
  Dongxuan Liu\textsuperscript{1}\quad
  Yibin Lou\textsuperscript{3}\quad
  Jie Wu\textsuperscript{2}\quad
  Jing Xu\textsuperscript{7}\\
  Jian Zhang\textsuperscript{1}\quad
  Zheng Yang\textsuperscript{1}\quad
  Yunlong Lin\textsuperscript{1}\quad
  Chenxin Li\textsuperscript{4}\quad
  Panwang Pan\textsuperscript{1}\quad
  Junbin Lu\textsuperscript{5}\quad
  Jingyan Jiang\textsuperscript{6}\\
  Xinghao Ding\textsuperscript{1}\quad
  Yue Huang\textsuperscript{\dag\,1}\quad
  Zhi Wang\textsuperscript{2}\\[2ex]
  \textsuperscript{1}XMU\quad
  \textsuperscript{2}THU\quad
  \textsuperscript{3}SUSTech\quad
  \textsuperscript{4}CUHK\quad
  \textsuperscript{5}UW\quad
  \textsuperscript{6}SZTU\quad
  \textsuperscript{7}JNU\\[1ex]
  {\small *\,Equal contributions.\quad \dag\,Corresponding author.\quad \spademark\,Project lead.}
}

\usepackage{etoolbox}

\makeatletter
\patchcmd{\abstract}{\vspace{0pt}}{\vspace{0pt}}{}{}   % ← 把 6pt 改成你想要的
\makeatother

\begin{document}

\twocolumn[{%
    \maketitle
    \vspace*{-0.4in}
    % \vspace{-1em} % 调整标题和 Teaser 图之间的间距
    \centering
        
    \includegraphics[width=1\textwidth]{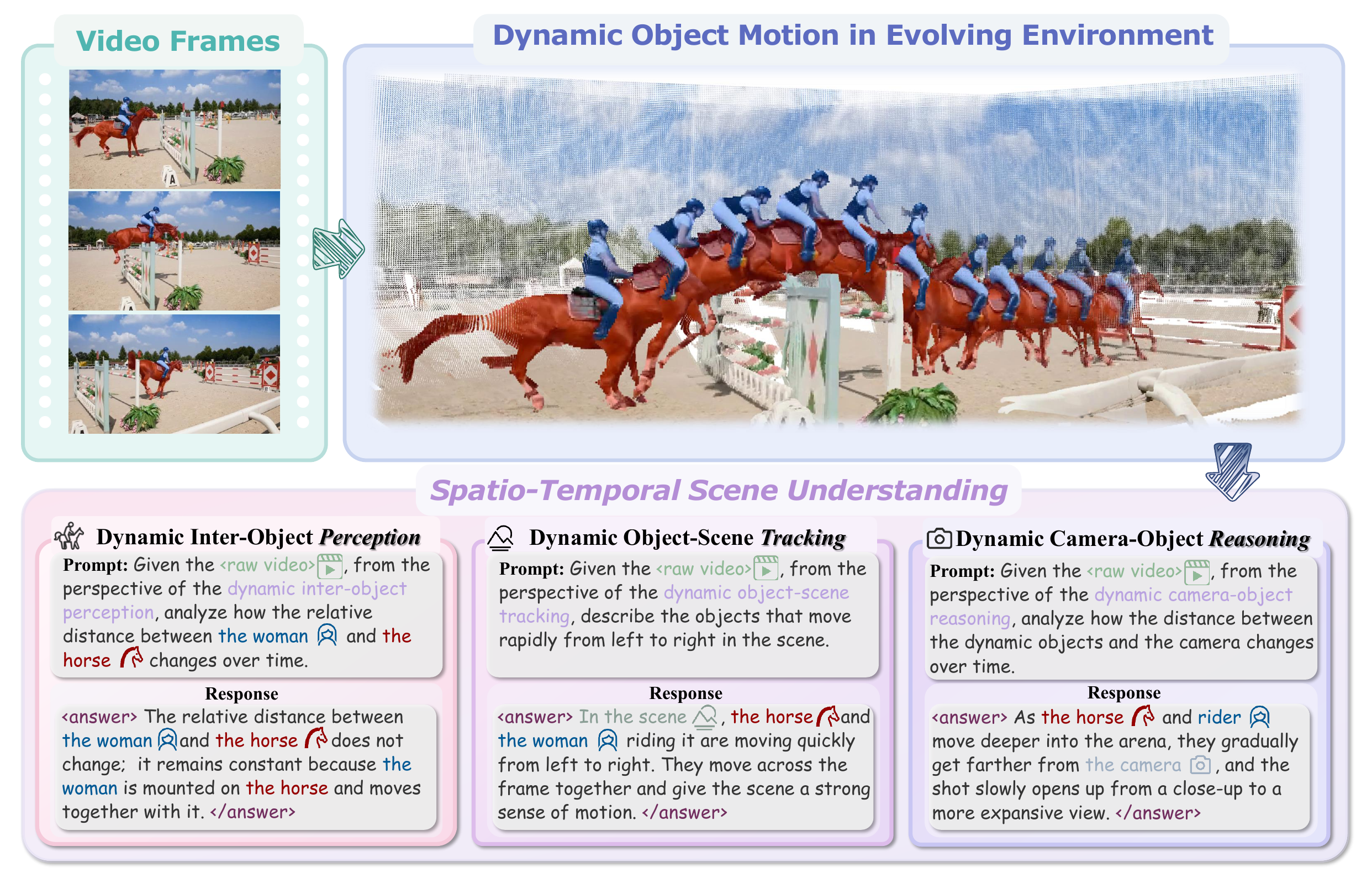}
    \vspace{-0.9cm}
    \captionof{figure}{\textbf{Spatio-temporal dynamics reasoning benchmark \texttt{Dyn-Bench}.}
    % \captionof{figure}{\textbf{Evaluation hierarchy of \texttt{Dyn-Bench} for spatio-temporal reasoning.}
    It rigorously evaluates multimodal large language models on their ability to \textit{perceive}, \textit{track}, and \textit{reason} about dynamic contents (\textit{i.e.}, object motion, evolving scene, camera motion) in the 4D world.}
    \label{fig:teaser}
    \vspace{0.1cm} % 调整 Teaser 图和摘要之间的间距
}]

\vspace{-10mm}
\begin{abstract}
Humans inhabit a physical 4D world where geometric structure and semantic content evolve over time, constituting a dynamic 4D reality (spatial with temporal dimension). While current Multimodal Large Language Models (MLLMs) excel in static visual understanding, can they also be adept at “thinking in dynamics”, \textit{i.e.}, perceive, track and reason about spatio-temporal dynamics in evolving scenes?
To systematically assess their spatio-temporal reasoning and localized dynamics perception capabilities, we introduce \texttt{Dyn-Bench}, a large-scale benchmark built from diverse real-world and synthetic video datasets, enabling robust and scalable evaluation of spatio-temporal understanding.
Through multi-stage filtering from massive 2D and 4D data sources, \texttt{Dyn-Bench} provides a high-quality collection of dynamic scenes, comprising 1k videos, 7k visual question answering (VQA) pairs, and 3k dynamic object grounding pairs. 
We probe general, spatial and region-level MLLMs to express how they think in dynamics both linguistically and visually, and find that existing models cannot simultaneously maintain strong performance in both spatio-temporal reasoning and dynamic object grounding, often producing inconsistent interpretations of motion and interaction. 
Notably, conventional prompting strategies (\textit{e.g.}, chain-of-thought or caption-based hints) provide limited improvement, whereas structured integration approaches, including Mask-Guided Fusion and Spatio-Temporal Textual Cognitive Map (ST-TCM), significantly enhance MLLMs' dynamics perception and spatio-temporal reasoning in the physical 4D world. Code and benchmark are available at \url{https://dyn-bench.github.io/}.
\end{abstract}

\vspace{-0.5cm}
\vspace{-2mm}
\section{Introduction}
\label{sec:intro}
\vspace{-1mm}

% -- Org --
% Humans live in a continuously evolving 4D physical world where spatial structures and semantic content change over time.
% By perceiving, tracking, and reasoning about dynamic objects, human cognition naturally integrates motion, interaction, and temporal change, a process we call \textit{thinking in dynamics}.
% This raises a fundamental question: \textit{How Multimodal Large Language Models perceive, track, and reason about the physical 4D world like humans?}

Understanding the dynamics of 4D world, where both spatial structure and semantic content evolve over time, is fundamental to how humans interact with environment. This dynamic perception allows humans to reason about motion, interactions, and temporal changes, seamlessly integrating visual input with temporal context. While humans rely on visual intelligence to navigate and interpret static scenes, their ability to \textit{think in dynamics}, by tracking and reasoning about evolving scenes through the connections between objects' movements, interactions, and spatial relationships over time, is what enables them to understand dynamic environments. Although current MLLMs excel in spatial reasoning, their ability to understand spatio-temporal dynamics in evolving scenes remains underexplored. This motivates us to ask: \textit{To what extent can MLLMs perceive, track, and reason about the physical 4D world in a similarly coherent manner as humans?}

%物理世界是4d世界，人类能够能够通过对实例进行感知来进行时空推理
Recent advances in general MLLMs~\cite{Gemini2.5, gpt-5, Gpt-4o, Qwen2.5-VL, Qwen3-Omni, InternVL3.5, LLaVA-OneVision-1.5, GLM-4.5V} and spatial reasoning variants~\cite{spacer, spatialcot, spatialmllm, scener1, 3dr1, sr3d, spatailvlm, vlm3r, vst, spatial-ssrl} have significantly advanced visual intelligence.
While these models demonstrate strong semantic and spatial reasoning capabilities on static images, their success does not readily generalize to dynamic video understanding.
When applied to temporally evolving scenes, they struggle to maintain consistent representations of moving objects across frames due to challenges such as occlusion, ego-motion, and semantic ambiguity.
These limitations often lead to a fragmented understanding of motion and interactions, hindering coherent reasoning in dynamic environments.
To address these issues, recent \textit{region-level} MLLMs~\cite{Sa2va, Unipixel, PerceiveAnything, Describeanything, pixelrefer, Videorefer, VideoGLaMM, GraspAnyRegion, Vitron} pursue fine-grained, object-centric understanding through explicit region–language alignment, improving localized spatial perception and tracking accuracy.
However, such progress remains largely confined to \textit{spatial reasoning}, without extending to the broader challenge of \textit{spatio-temporal reasoning}, which requires continuous perception, tracking, and reasoning about dynamic objects over time.

% To comprehensively evaluate spatio-temporal reasoning, recent benchmarks~\cite{DSI-Bench,VSI-bench,Sti-bench,Ost-bench,ODI-Bench,Vlm4d} have extended evaluation beyond static visual understanding toward dynamic scene reasoning.

% Recent benchmarks~\cite{DSI-Bench, VSI-bench, Sti-bench, Ost-bench, ODI-Bench, Vlm4d} have extended the evaluation of spatio-temporal reasoning beyond static visual understanding to dynamic scene reasoning.
% However, as summarized in Table~\ref{tab:benchmark_comparison}, most existing efforts remain focused on scene-level modeling, emphasizing temporal correlations, motion understanding, or scene evolution, while lacking systematic evaluation from the perspective of \textit{dynamic objects}.
% As a result, these benchmarks fail to capture the multi-dimensional nature of spatio-temporal reasoning, particularly the ability to continuously track moving entities and delineate their fine-grained motion boundaries over time.
% Without explicitly modeling such object-centric dynamics, current evaluations fall short in assessing whether MLLMs can achieve coherent reasoning about motion, causality, and scene evolution in realistic 4D environments.

Recent benchmarks~\cite{DSI-Bench, VSI-bench, Sti-bench, Ost-bench, ODI-Bench, Vlm4d} have expanded the evaluation of spatio-temporal reasoning beyond static visual understanding to dynamic scene reasoning. 
However, as summarized in Tab.~\ref{tab:benchmark_comparison}, most existing efforts focus on scene-level modeling, emphasizing temporal correlations, motion understanding, or scene evolution, but lack systematic evaluation from the perspective of \textit{dynamic objects}. Consequently, these benchmarks fail to capture the multi-dimensional nature of spatio-temporal reasoning, particularly the ability to track moving objects and delineate their  fine-grained motion boundaries over time. Without modeling object-centric dynamics, current evaluations fall short in assessing whether MLLMs can achieve coherent reasoning about motion, causality, and scene evolution in realistic 4D environments.

\vspace{-0.5mm}
% (\textbf{Dyn}amic-object \textbf{Bench}mark)
To this end, we introduce \texttt{Dyn-Bench} (\textbf{Dyn}amics \textbf{Bench}mark), a large-scale benchmark for evaluating the \textit{spatio-temporal reasoning} and \textit{dynamic object grounding} abilities of MLLMs in the physical 4D world.
As illustrated in Fig.~\ref{fig:teaser}, \texttt{Dyn-Bench} spans three complementary levels of dynamic scene understanding:  
\ding{172}~\textit{Dynamic Inter-Object Perception}, capturing spatial relations and interactions among moving objects;  
\ding{173}~\textit{Dynamic Object–Scene Tracking}, modeling object motion and temporal evolution across scenes; and  
\ding{174}~\textit{Dynamic Camera–Object Reasoning}, analyzing object behavior under varying camera motions.  
Each level integrates spatio-temporal reasoning and dynamic object grounding tasks for joint evaluation of perception and reasoning.  
Constructed from eight real and synthetic datasets, \texttt{Dyn-Bench} contains 1k videos, 7k visual question answering (VQA) pairs, and 3k grounding annotations.  
We extensively evaluate general, spatial and region-level MLLMs, finding that existing models struggle to balance reasoning and grounding, often showing inconsistent understanding of object motion over time.  
Building on these findings, we further explore how MLLMs \textit{think in dynamics} across visual and textual modalities.  
Our analysis reveals that conventional prompting strategies, such as chain-of-thought or caption-based cues, yield limited gains, whereas structured integration methods, including Mask-Guided Fusion and the proposed ST-TCM, enable more coherent modeling of motion and significantly improve spatio-temporal reasoning on \texttt{Dyn-Bench}.

In summary, our contributions are as follows:
\begin{itemize}
\item[$\bullet$] We introduce \texttt{Dyn-Bench}, the first benchmark that comprehensively evaluates Dynamic Understanding capability of MLLMs, including spatio-temporal reasoning and dynamic object grounding in realistic 4D environments.

\item[$\bullet$] We introduce a novel, carefully curated dataset, constructed from eight real and synthetic sources. After multi-stage filtering from extensive 2D and 4D data, the dataset includes 1k videos, 7k VQA pairs, and 3k high-quality grounding annotations.

\item[$\bullet$] We analyze the limitations of current MLLMs in spatio-temporal reasoning, identifying issues such as inconsistent object motion tracking and poor grounding of dynamic objects. We also evaluate the impact of different prompting techniques, including chain-of-thought reasoning and mask-guided methods, demonstrating that structured approaches enhance the coherence of reasoning in dynamic scenes.

\item[$\bullet$]  We introduce Spatio-Temporal Textual Cognitive Maps (ST-TCM), a novel framework that unifies spatial, temporal, and motion information into a single representation. ST-TCM significantly enhances the ability of MLLMs to reason about dynamic objects and scenes, leading to more consistent and accurate spatio-temporal understanding.
\end{itemize}

\vspace{-2mm}
\begin{table}[t]
    \centering
    \caption{\textbf{Comparison of \texttt{Dyn-Bench} with existing spatio-temporal benchmarks.} \texttt{Dyn-Bench} offers a unified assessment covering three levels of dynamic object grounding and reasoning while spanning more diverse data domains than prior works.}
    \vspace{-3mm}
    \resizebox{0.98\linewidth}{!}{ % 更小的缩放比例
    \begin{tabular}{lccccccccccccccc}
    \toprule
    \multicolumn{1}{c}{} & \multicolumn{3}{c}{\cellcolor[HTML]{E2ECEF}\textit{\textcolor[HTML]{00536b}{Numerical Statistics}}} & \multicolumn{4}{c}{\cellcolor[HTML]{E1E8F3}\textit{\textcolor[HTML]{324779}{QA Types}}} & \multicolumn{4}{c}{\cellcolor[HTML]{E5E1F5}\textit{\textcolor[HTML]{5a3477}{Detailed Features}}} \\
    \multicolumn{1}{c}{\multirow{-2}{*}{Benchmark}} &
    \cellcolor[HTML]{E2ECEF}\rotatebox{90}{\makecell{\textcolor[HTML]{00536b}{\# Videos}}} &
    \cellcolor[HTML]{E2ECEF}\rotatebox{90}{\makecell{\textcolor[HTML]{00536b}{\# QA Pairs}}} &
    \cellcolor[HTML]{E2ECEF}\rotatebox{90}{\makecell{\textcolor[HTML]{00536b}{\# Masklets}}} &
    \cellcolor[HTML]{E1E8F3}\rotatebox{90}{\makecell{\textcolor[HTML]{324779}{Inter-Object}}} &
    \cellcolor[HTML]{E1E8F3}\rotatebox{90}{\makecell{\textcolor[HTML]{324779}{Object-Scene}}} &
    \cellcolor[HTML]{E1E8F3}\rotatebox{90}{\makecell{\textcolor[HTML]{324779}{Camera-Object}}} &
    \cellcolor[HTML]{E1E8F3}\rotatebox{90}{\makecell{\textcolor[HTML]{324779}{Object Grounding}}} &
    \cellcolor[HTML]{E5E1F5}\rotatebox{90}{\makecell{\textcolor[HTML]{5a3477}{Scene Type}}} &
    \cellcolor[HTML]{E5E1F5}\rotatebox{90}{\makecell{\textcolor[HTML]{5a3477}{Dynamic Type}}} &
    \cellcolor[HTML]{E5E1F5}\rotatebox{90}{\makecell{\textcolor[HTML]{5a3477}{Real-world?}}} &
    \cellcolor[HTML]{E5E1F5}\rotatebox{90}{\makecell{\textcolor[HTML]{5a3477}{Metric-scale?}}} \\ 
    \hline
    EgoDynamic4D~\cite{UnderstandingDynamicEgoCentric4D} & 275 & 927,000 & - &  \usym{2713} & \usym{2713} & \usym{2713} & \usym{2717}  & Indoor & S.fisheye & Mixed & Yes\\
    Chat4D~\cite{LLaVA-4D} & - & 879,100 & - &   \usym{2717} &  \usym{2713} & \usym{2717} & \usym{2717} & Mixed & Realistic & Reak-world & Yes \\
    DynSuperCLEVR~\cite{compositional} & 1200 & 11,589 & -  & \usym{2713} &  \usym{2713} & \usym{2717} & \usym{2717}  & Outdoor & Open-field & Synthetic & Yes \\
    VideoSTR~\cite{VideoSTR} & - & 205,000 & - & \usym{2713} & \usym{2713} &  \usym{2717} &  \usym{2717} & Mixed &  Realistc & Real-world  & Yes \\
    STI-Bench~\cite{Sti-bench} & 300 & 2,000 & -  & \usym{2713} & \usym{2713} & \usym{2713}  &  \usym{2717} & Mixed & Driving  & Real-world & Yes \\
    OST-Bench~\cite{Ost-bench} & 1,400 & 10,000 & - & \usym{2717} & \usym{2717} & \usym{2713} & \usym{2717}& Indoor & Room  & Real-world  & Yes \\
    VLM4D~\cite{Vlm4d} &  1,000 & 1,816 & - & \usym{2713} & \usym{2713} & \usym{2713} & \usym{2717} & Outdoor & Realistic  &  Mixed &  No \\
    DSI-Bench~\cite{DSI-Bench} & 943 & 1,700 & -   & \usym{2717} & \usym{2717} & \usym{2713} & \usym{2717} & Outdoor &  Realistic &  Real-world & No \\

    \midrule
    \textbf{Dyn-Bench} & 1000 & 7,000 & 3,000 & \usym{2713} & \usym{2713} & \usym{2713} &\usym{2713} & Mixed & Realistic &  Mixed & Yes \\
    \bottomrule
    \end{tabular}
    } %        
    \label{tab:benchmark_comparison}
    \vspace{-3mm}
\end{table}

\begin{figure*}[t]  % 使用 figure* 确保在顶部
    \centering
    \includegraphics[width=1\linewidth]{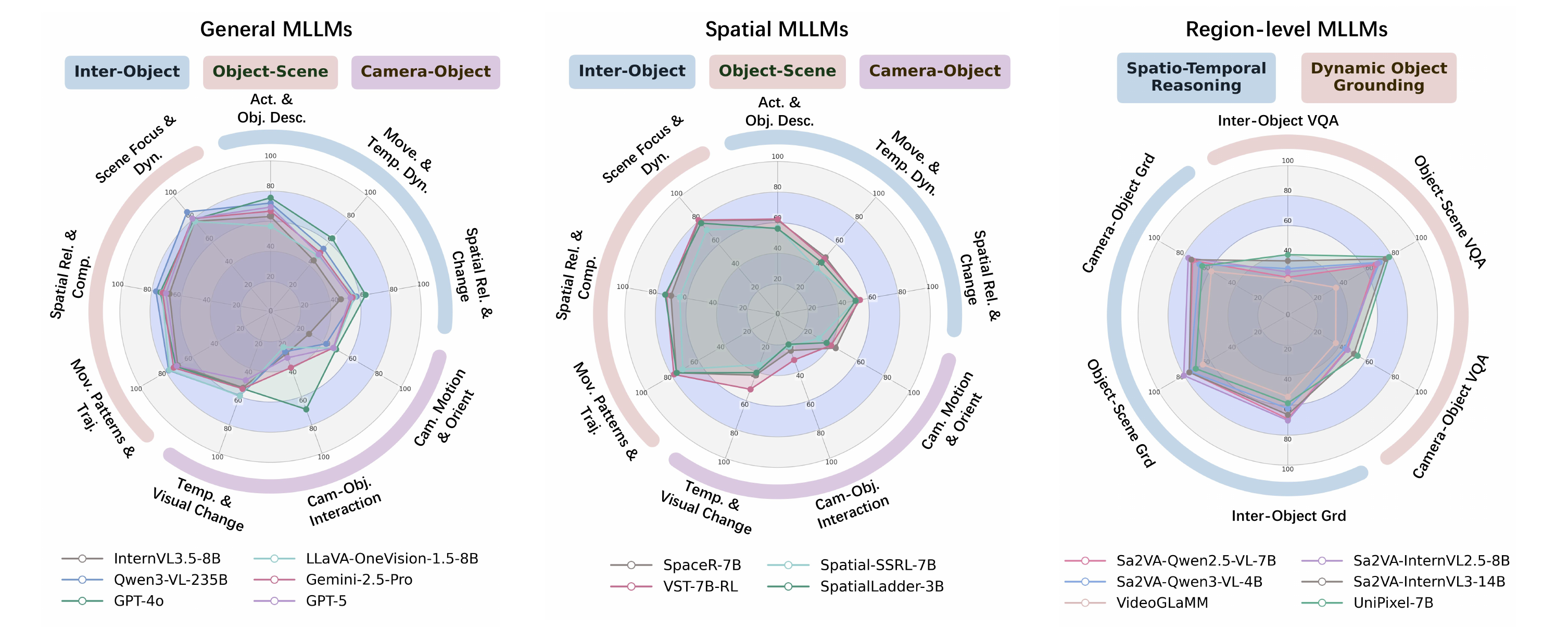}  % 保留第一个图
    \vspace{-10mm}
    \caption{\textbf{Model performance on \texttt{Dyn-Bench}.}
    \textbf{Left} / \textbf{Center} radar charts show general and spatial MLLMs accuracy on nine spatio-temporal tasks; \textbf{Right} radar chart shows region-level MLLMs performance on spatio-temporal reasoning and dynamic object grounding.}
    \label{fig:mllms_compare}
    \vspace{-3mm}  % 控制图像下方的间距
\end{figure*}

\vspace{-1mm}
\section{\texttt{Dyn-Bench}}

\label{sec:dyn-bench}
%缺少一个每个level详细的可视化图，具体构造的vqa和mask question 是什么样的

\vspace{-1mm}

\begin{figure*}[ht]
    \centering
    \includegraphics[width=1\textwidth]{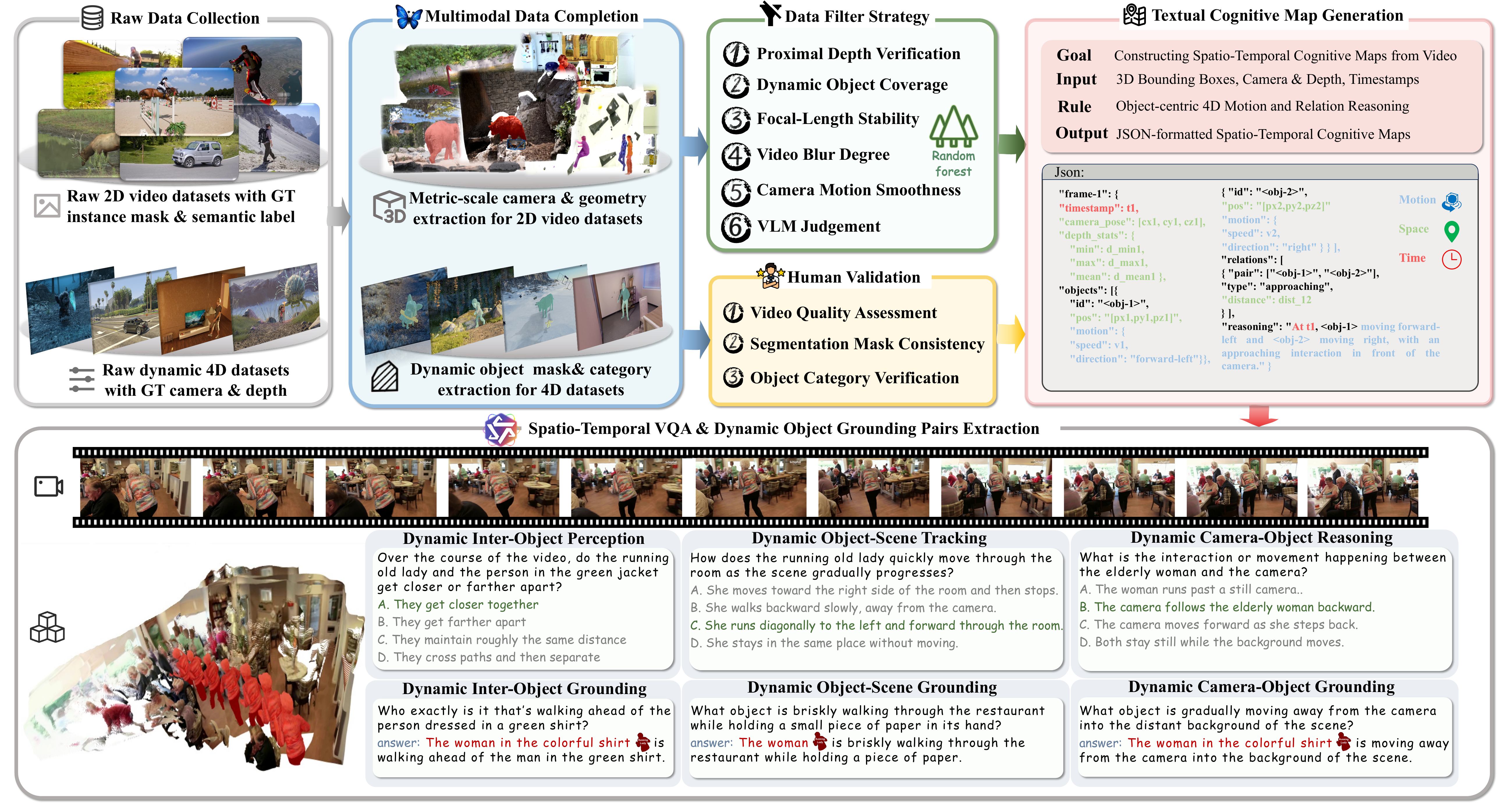} % 图片名要和上传一致
    \caption{\textbf{Benchmark curation pipeline.} The pipeline integrates dynamic video datasets from multiple sources, followed by multimodal completion with geometry and mask extraction. Data filter strategy ensures consistency and motion stability, complemented by human validation for quality assurance. Finally, spatial-temporal VQA and grounding pairs are generated with structured textual cognitive maps.}
    \label{fig:data_engine}
        \vspace{-5mm}
\end{figure*}
\vspace{-1mm}
\subsection{Overview}
\begin{figure}[h]  % b = bottom
    \centering      % 或去掉改成 \raggedright 让它靠左
    \includegraphics[width=0.9\linewidth]{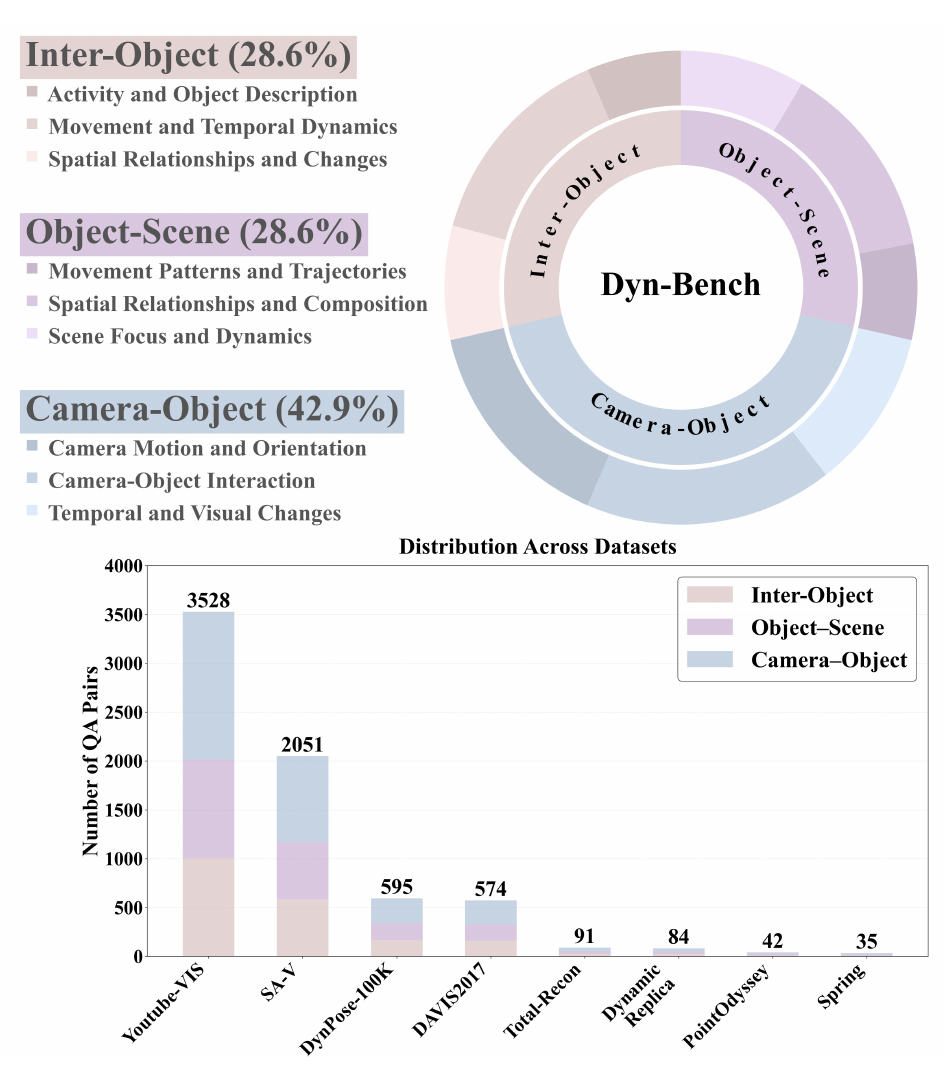}
    \caption{\textbf{Benchmark Statistics.} \textbf{Top}: Distribution of tasks across three levels. \textbf{Bottom}: VQA pairs distribution across datasets.}
    \label{fig:stastics}
    \vspace{-9mm}
\end{figure}

\vspace{-1mm}
We present \texttt{Dyn-Bench}, a large-scale benchmark for quantitatively evaluating the spatio-temporal reasoning abilities of MLLMs under fine-grained dynamic-object understanding.
\texttt{Dyn-Bench} consists of 1k dynamic video scenes with 7k visual question answering pairs and 3k grounding annotations, collected from four 2D video segmentation and four 4D dynamic scene datasets spanning diverse environments, motion patterns, and camera trajectories.
As illustrated in Fig.~\ref{fig:data_engine}, the benchmark is structured into three complementary levels: \textit{Dynamic Inter-Object Perception}, \textit{Dynamic Object–Scene Tracking}, and \textit{Dynamic Camera–Object Reasoning}, each integrating spatio-temporal reasoning and dynamic object grounding tasks.
An overview of dataset statistics is provided in Fig.~\ref{fig:stastics}.

\vspace{-2mm}

\subsection{Benchmark Construction}
\vspace{-1mm}

\noindent \textbf{Data Collection and Filtering.}  
We construct \texttt{Dyn-Bench} by collecting dynamic videos from four 2D video segmentation datasets (DAVIS~\cite{Davis}, SA-V~\cite{Sa-V}, DynPose-100K~\cite{DynPose-100K}, and YouTube-VIS~\cite{YouTube-VIS}) and four 4D dynamic-scene datasets (DynamicReplica~\cite{DynamicReplica}, PointOdyssey~\cite{Pointodyssey}, Spring~\cite{Spring}, and Total-Recon~\cite{totalrecon}). These datasets provide instance masks, depth maps, and camera poses, enabling accurate question–answer generation and object category annotation. Missing annotations are completed using existing pipelines~\cite{Sa2va, vipe, qwen3technicalreport} to ensure cross-modal consistency.
To ensure data reliability, we employ a multi-criteria data filter strategy assessing geometric stability, motion smoothness, image sharpness, and depth consistency, supported by VLM-based quality evaluation. Low-quality videos are discarded to maintain visual and geometric fidelity. Filtering statistics are shown in Tab.~\ref{tab:data_filtering}, with additional details provided in the supplementary material.

\begin{table}[h]
\centering
\caption{\textbf{Dataset statistics across \texttt{Dyn-Bench} filtering stages.}}
\vspace{-1mm}
\small
\setlength{\tabcolsep}{3pt}
\renewcommand{\arraystretch}{0.9}
\resizebox{0.85\linewidth}{!}{
\begin{tabular}{lccc}
\toprule
\textbf{Dataset} & 
\cellcolor[HTML]{E2ECEF}\textit{\textcolor[HTML]{00536b}{Original Num}} & 
\cellcolor[HTML]{E1E8F3}\textit{\textcolor[HTML]{324779}{Filtered Num}} & 
\cellcolor[HTML]{E5E1F5}\textit{\textcolor[HTML]{5a3477}{Human Selected}} \\
\midrule
\rowcolor{navyblue!5}
\multicolumn{1}{l!}{\textbf{\textit{2D Dataset}}} & & & \\
DAVIS~\cite{Davis}            & 200 & 89 & 82 \\
SA-V~\cite{Sa-V}              & 31,000 & 972 & 293 \\
DynPose-100k~\cite{DynPose-100K} & 3,888 & 199 & 85 \\
YouTube-VIS~\cite{YouTube-VIS}  & 2,981 & 666 & 504 \\
\rowcolor{navyblue!5}
\multicolumn{1}{l!}{\textbf{\textit{4D Dataset}}} & & & \\
DynamicReplica~\cite{DynamicReplica} & 224 & 36 & 12 \\
PointOdyssey~\cite{Pointodyssey} & 145 & 26 & 6 \\
Spring~\cite{Spring} & 43 & 14 & 5 \\
Total-Recon~\cite{totalrecon} & 26 & 24 & 13 \\
\midrule
\textbf{Total} & \textbf{38,507} & \textbf{2,026} & \textbf{1,000} \\
\bottomrule
\end{tabular}
}
\vspace{-2mm}
\label{tab:data_filtering}
\end{table}

    % 1st: \cellcolor[HTML]{FFB366}{1st}
    % 2nd: \cellcolor[HTML]{FFCC99}{2nd}
    % 3rd: \cellcolor[HTML]{FFE5CC}{3rd}

\begin{table*}[t]

   \caption{\textbf{Spatio-temporal reasoning evaluation on \texttt{Dyn-Bench}.}
    Top three performers in each task category are highlighted from 
    \colorbox[HTML]{FFB366}{Dark} (highest) to \colorbox[HTML]{FFE5CC}{Light} (third highest), 
    and overall model rankings are ranging from 
    \colorbox{oai-green-600}{Dark} (highest) to \colorbox{oai-green-200}{Light} (third highest).}

    \centering
    \fontsize{5pt}{5pt}\selectfont
    \setlength\tabcolsep{3pt}
    \renewcommand{\arraystretch}{1.20}
    \resizebox{0.9\textwidth}{!}{%
    \begin{tabular}{r!{\vrule width 0.9pt}cc!{\vrule width 0.9pt}ccccccccc}
        & & &
        \rotatebox[origin=c]{75}{Act. \& Obj. Desc.} &
        \rotatebox[origin=c]{75}{Move. \& Temp. Dyn.} &
        \rotatebox[origin=c]{75}{Spatial Rel. \& Change} &
        \rotatebox[origin=c]{75}{Mov. Patterns \& Traj.} &
        \rotatebox[origin=c]{75}{Spatial Rel. \& Comp.} &
        \rotatebox[origin=c]{75}{Scene Focus \& Dyn.} &
        \rotatebox[origin=c]{75}{Cam. Motion \& Orient.} &
        \rotatebox[origin=c]{75}{Cam-Obj. Interaction} &
        \rotatebox[origin=c]{75}{Temp. \& Visual Change} \\
        Methods & Rank & Avg. &
        \multicolumn{3}{c}{\cellcolor{orange!15}Inter-Object } &
        \multicolumn{3}{c}{\cellcolor{orange!10}Object-Scene } &
        \multicolumn{3}{c}{\cellcolor{yellow!10}Camera-Object} \\
        \noalign{\hrule height 0.9pt}

        \rowcolor{navyblue!5}
        \multicolumn{1}{l!{\vrule width 0.9pt}}{\textbf{\textit{Baseline}}} & & & & & & & & & & & \\
        Chance Level (Random) & - &  & 25.0 & 25.0 & 25.0 & 25.0 & 25.0 & 25.0 & 25.0 & 25.0 & 25.0 \\
        Chance Level (Frequency) & - &  & 12.3 & 21.6 & 31.5 & 29.1 & 27.7 & 13.8 & 33.5 & 10.1 & 25.6 \\
        \noalign{\hrule height 0.6pt}

        \rowcolor{navyblue!5}
        \multicolumn{1}{l!{\vrule width 0.9pt}}{\textbf{\textit{Proprietary Models (API)}}} & & & & & & & & & & & \\
        GPT-4o & \cellcolor{oai-green-200}{3} &
        50.1 & 56.1 & 38.7 & 44.6 &
        63.1 & 59.1 & 68.8 & 47.2 & 42.0 & 49.2 \\
        GPT-5 & \cellcolor{oai-green-400}{2} & 59.5 & 68.6 &
        47.3 & 48.1 & 71.7 & 65.9 & 73.0 &
        \cellcolor[HTML]{FFB366}{60.9} &
        \cellcolor[HTML]{FFCC99}{58.4} &
        \cellcolor[HTML]{FFE5CC}{58.4} \\
        Gemini-2.5 Pro & \cellcolor{oai-green-600}{1} &
        59.8 & 69.7 & 48.0 &
        50.5 & 67.8 & 59.9 & 65.6 &
        \cellcolor[HTML]{FFCC99}{60.7} &
        \cellcolor[HTML]{FFE5CC}{54.9} & 51.8 \\
        \noalign{\hrule height 0.6pt}

        \rowcolor{navyblue!5}
        \multicolumn{1}{l!{\vrule width 0.9pt}}{\textbf{\textit{Open-source Models}}} & & & & & & & & & & & \\
        InternVL3-14B & 7 & 53.7 & 65.3 & 47.0 & 49.7 & 67.8 & 69.2 & 77.7 & 37.9 & 44.9 & 46.9 \\
        InternVL3-38B & 5 & 54.2 & 68.2 & 44.8 & 48.4 & 71.1 & 67.3 & 76.7 & 41.2 & 48.9 & 44.6 \\
        InternVL3.5-8B & 11 & 50.3 & 66.6 & 41.0 & 44.8 & 63.5 & 62.7 & 69.6 & 41.5 & 40.1 & 46.4 \\
        InternVL3.5-38B & 10 & 50.8 & 65.0 & 40.7 & 42.4 & 60.2 & 62.3 & 72.3 & 45.5 & 42.9 & 44.6 \\
        Qwen2.5-VL-7B & 9 & 51.6 & 61.7 & 42.8 & 48.0 & 69.1 & 67.0 & 73.7 & 43.8 & 39.6 & 42.8 \\
        Qwen2.5-VL-32B & 4 & 56.0 &
        \cellcolor[HTML]{FFE5CC}{71.5} &
        52.2 & \cellcolor[HTML]{FFE5CC}{53.8} & 71.5 & 67.6 & 75.3 & 42.1 & 46.7 & 47.9 \\
        Qwen2.5-VL-72B & 8 & 51.8 & 65.5 & 41.1 & 43.7 & 60.4 & 57.8 & 68.3 & 49.2 & 42.6 & 55.0 \\
        Qwen3-VL-8B & \cellcolor{oai-green-200}{3} & 61.4 &
        70.8 & \cellcolor[HTML]{FFE5CC}{52.6} &
        53.6 & \cellcolor[HTML]{FFE5CC}{75.0} &
        \cellcolor[HTML]{FFE5CC}{71.2} &
        \cellcolor[HTML]{FFCC99}{82.4} &
        55.4 & 52.6 &
        \cellcolor[HTML]{FFCC99}{60.0} \\
        Qwen3-VL-32B & \cellcolor{oai-green-400}{2} & 62.7 &
        \cellcolor[HTML]{FFCC99}{73.7} &
        \cellcolor[HTML]{FFB366}{56.2} &
        53.4 & 74.6 & \cellcolor[HTML]{FFCC99}{73.1} &
        \cellcolor[HTML]{FFE5CC}{80.2} &
        58.2 & 54.3 & 56.9 \\
        Qwen3-VL-235B & \cellcolor{oai-green-600}{1} & 65.3 &
        \cellcolor[HTML]{FFB366}{76.4} &
        \cellcolor[HTML]{FFCC99}{55.8} &
        \cellcolor[HTML]{FFB366}{55.6} &
        \cellcolor[HTML]{FFB366}{77.8} &
        \cellcolor[HTML]{FFB366}{76.1} &
        \cellcolor[HTML]{FFB366}{84.1} &
        \cellcolor[HTML]{FFE5CC}{59.8} &
        \cellcolor[HTML]{FFB366}{59.0} &
        \cellcolor[HTML]{FFB366}{60.2} \\
        LLaVA-OneVision-1.5-4B & 12 & 49.9 & 50.5 & 48.5 & 50.3 & 65.2 & 64.8 & 63.9 & 39.9 & 36.6 & 46.1 \\
        LLaVA-OneVision-1.5-8B & 6 & 53.8 &
        60.9 & 47.7 & 53.4 &
        74.4 & 69.6 & 75.4 & 41.0 & 37.0 & 51.6 \\
        \rowcolor{navyblue!5}
        \multicolumn{1}{l!{\vrule width 0.9pt}}{\textbf{\textit{Spatial MLLMs}}} & & & & & & & & & & & \\
          SpaceR-7B&\cellcolor{oai-green-600}{1} & 56.5 & 66.6 & 49.2 & 52.7 & 72.2 & 67.8 & 78.2 & 50.3 & 40.0 & 55.5 \\
          VST-7B-RL& \cellcolor{oai-green-400}{2} & 55.7 & 68.6 & 48.4 & 51.9 &
          73.0 & 70.7 & 79.4 & 45.1 & 39.1 & 52.9 \\
          Spatial-SSRL-7B& 4 & 45.9 & 54.5 & 40.0 & 48.1 & 68.5 & 65.9 & 73.8 & 35.8 & 36.7 & 37.7 \\
         SpatialLadder-3B& \cellcolor{oai-green-200}{3} & 53.6 & 60.8 & 46.1 & 49.2 & 70.0 & 70.9 & 77.1 & 38.2 & 42.0 & 51.9 \\
        \rowcolor{navyblue!5}
        \multicolumn{1}{l!{\vrule width 0.9pt}}{\textbf{\textit{Region-level MLLMs}}} & & & & & & & & & & & \\
        UniPixel-3B & \cellcolor{oai-green-400}{2} & 55.4 &
        63.3 & 47.2 & 53.2 & 71.7 & 70.2 &
        77.7 & 43.2 & 43.6 & 52.0 \\
        UniPixel-7B & \cellcolor{oai-green-600}{1} & 58.1 &
        64.4 & 50.2 & \cellcolor[HTML]{FFCC99}{54.7} &
        \cellcolor[HTML]{FFCC99}{76.1} & 70.4 & 79.7 & 47.3 &
        47.3 & 55.7 \\
        VideoGLaMM & 7 & 30.7 & 35.6 & 34.4 & 35.0 & 34.6 & 38.2 & 39.3 & 22.7 & 21.2 & 25.9 \\
        Sa2VA-InternVL2.5-8B & 6 & 49.4 & 61.0 & 42.4 & 45.7 & 66.1 & 62.8 & 71.9 & 36.6 & 36.4 & 47.2 \\
        Sa2VA-InternVL3-14B & \cellcolor{oai-green-200}{3} & 53.6 & 55.9 & 48.9 & 53.2 & 72.0 & 70.2 & 74.6 & 38.1 & 39.6 & 53.6 \\
        Sa2VA-Qwen2.5-VL-7B & 4 & 50.3 & 58.6 & 39.3 & 52.9 & 67.6 & 62.1 & 70.5 & 38.8 & 39.1 & 49.3 \\
        Sa2VA-Qwen3-VL-4B & 5 & 49.8 & 60.8 & 39.3 & 46.2 & 67.2 & 62.0 & 73.2 & 41.0 & 44.5 & 36.8 \\
    \end{tabular}
    }% end resizebox
 
    \vspace{-5mm}
    \label{tab:dynbench_table}
\end{table*}

\noindent \textbf{Question-Answer Generation.}
Based on the filtered video collection, we employ a ST-TCM in conjunction with Qwen3-VL~\cite{Qwen3-Omni} to construct dynamic-object-centered VQA tasks. The benchmark evaluates MLLMs across three complementary dimensions: \textit{Dynamic Inter-Object Perception}, focusing on how models perceive and interpret motion interactions and spatial relations among multiple dynamic objects (\textit{e.g.}, approach, occlusion, or overtaking); \textit{Dynamic Object-Scene Tracking}, capturing how individual objects are temporally tracked and evolve within continuously changing scenes (\textit{e.g.}, entering, leaving, or undergoing functional transitions); and \textit{Dynamic Camera-Object Reasoning}, assessing how camera motion influences the perceived geometry, depth, and temporal consistency of dynamic objects (\textit{e.g.}, relative translation, rotation, or event order). Each VQA dimension is paired with a corresponding object grounding task that associates the referenced dynamic objects with their instance segmentation masks. Dimension-specific prompting strategies and ST-TCM configurations are applied to Qwen3-VL to ensure focused spatio-temporal reasoning.

\noindent \textbf{Spatio-Temporal Textual Cognitive Map Construction.}
To capture fine-grained object motion and interactions in dynamic scenes, we construct a \textit{Spatio-Temporal Textual Cognitive Map (ST-TCM)} for each filtered video.
Given per-frame RGB-D inputs and segmentation masks, 3D object trajectories are reconstructed to obtain geometric attributes such as position, size, and orientation in world coordinates.
We then model inter-object and camera–object relations based on spatial proximity and motion continuity, capturing dynamic behaviors such as interaction and relative movement.
All geometric and spatial cues are translated into textual descriptions through a rule-based template system, integrating object geometry, motion, and relational dynamics into a unified spatio-temporal representation.
This structured textual form serves as input to \textit{Qwen3-VL-235B}~\cite{Qwen3-Omni} for dynamic object centered visual question answering and grounding.
Detailed implementation procedures are provided in the supplementary material.
\vspace{-2mm}

\noindent \textbf{Human Quality Control.}  
To ensure the reliability and perceptual validity of the filtered videos and generated annotations, we conduct an additional round of human verification covering video quality, mask consistency, VQA accuracy, and dynamic object category identification.  
Annotators assess camera stability, motion smoothness, and scene complexity to confirm visual quality, and examine segmentation masks to verify temporal coherence and consistent object identity across frames.  
The generated VQA and grounding pairs are also reviewed to ensure accurate object reference, reasoning-level alignment, and consistency with visual evidence.  
A summary of the multi-stage filtering and verification process is presented in Tab.~\ref{tab:data_filtering}.

\vspace{-2mm}
\section{Evaluation on \texttt{Dyn-Bench}}
\label{sec:evaluation}
\vspace{-1.5mm}
\subsection{Evaluation Setup}
\vspace{-1.5mm}
\noindent \textbf{Benchmark Models.}  
We evaluate three categories of MLLMs: general, spatial, and region-level models. General MLLMs (\textit{e.g.}, GPT-4o~\cite{Gpt-4o}, Qwen3-VL~\cite{Qwen3-Omni}) and spatial MLLMs (SpaceR~\cite{spacer}, VST~\cite{vst}, Spatial-SSRL~\cite{spatial-ssrl}, SpatialLadder~\cite{spatialladder}) lack explicit dynamic object grounding and are therefore evaluated only on spatio-temporal reasoning. In contrast, region-level MLLMs (Sa2VA~\cite{Sa2va}, UniPixel~\cite{Unipixel}, VideoGLaMM~\cite{VideoGLaMM}) are assessed on both spatio-temporal reasoning and dynamic object grounding to measure their fine-grained dynamic understanding.
To benchmark performance on spatio-temporal reasoning, we include two chance baselines: \textit{Chance Level (Random)}, obtained by uniformly sampling answers for multiple-choice questions, and \textit{Chance Level (Frequency)}, obtained by predicting the most frequent answer per task. All models are evaluated in a zero-shot setting using their default instruction templates to ensure consistent and fair comparison.

\noindent \textbf{Metric Design.}
For the three VQA task levels, we adopt a multiple-choice answering format and use \textit{Accuracy (ACC)} as the primary evaluation metric, following standard practice~\cite{Mmmu, Video-mme, Measuringmassive}. ACC is computed by exact matching over the model’s selected multiple-choice option.
For the corresponding object grounding tasks~\cite{Davis, mevis, visa}, we employ the video object segmentation metric $\mathcal{J\&F}$, which averages region similarity ($\mathcal{J}$) and boundary accuracy ($\mathcal{F}$).

\vspace{-1mm}

\begin{table*}[ht]

\caption{\textbf{Dynamic object grounding evaluation of Region-level MLLMs on \texttt{Dyn-Bench}.}
Top three performers in each task category are highlighted from 
\colorbox[HTML]{FFB366}{Dark} (highest) to \colorbox[HTML]{FFE5CC}{Light} (third highest).}

\centering
\setlength\tabcolsep{1.5pt}
\renewcommand{\arraystretch}{1.0}
\resizebox{0.8\textwidth}{!}{
\begin{tabularx}{\textwidth}{@{} r *{12}{Y} @{}}
\Xhline{1pt}
\multirow{2}{*}{\textbf{Models}} &
\multicolumn{3}{c}{\textbf{Average}} &
\multicolumn{3}{c}{\textbf{Inter-Object}} &
\multicolumn{3}{c}{\textbf{Object-Scene}} &
\multicolumn{3}{c}{\textbf{Camera-Object}} \\
\cmidrule(lr){2-4}\cmidrule(lr){5-7}\cmidrule(lr){8-10}\cmidrule(lr){11-13}
& $\mathcal{J}$ & $\mathcal{F}$ & $\mathcal{J\&F}$ &
  $\mathcal{J}$ & $\mathcal{F}$ & $\mathcal{J\&F}$ &
  $\mathcal{J}$ & $\mathcal{F}$ & $\mathcal{J\&F}$ &
  $\mathcal{J}$ & $\mathcal{F}$ & $\mathcal{J\&F}$ \\
\hline

\rowcolor{navyblue!5}
\multicolumn{1}{l!}{\textbf{\textit{Region-level MLLMs}}} & & & & & & & & & & & & \\

UniPixel-3B            
& 40.6 & 41.2 & 40.9
& 37.7 & 38.3 & 38.1
& 45.8 & 46.6 & 46.2
& 38.2 & 38.8 & 38.5 \\

UniPixel-7B            
& 64.4 & 66.0 & 65.2
& 65.4 & 66.6 & 66.0
& 70.1 & 72.0 & 71.1
& 57.8 & 59.4 & 58.6 \\

VideoGLaMM
& 55.4 & 63.8 & 59.6
& 54.8 & 63.0 & 58.9
& 61.4 & 69.8 & 65.6
& 49.9 & 58.7 & 54.3 \\

Sa2VA-InternVL2.5-8B
& \cellcolor[HTML]{FFB366}{74.2} & \cellcolor[HTML]{FFB366}{77.1} & \cellcolor[HTML]{FFB366}{75.6}
& \cellcolor[HTML]{FFB366}{75.4} & \cellcolor[HTML]{FFB366}{77.8} & \cellcolor[HTML]{FFB366}{76.8}
& \cellcolor[HTML]{FFB366}{78.6} & \cellcolor[HTML]{FFB366}{81.9} & \cellcolor[HTML]{FFB366}{80.2}
& \cellcolor[HTML]{FFB366}{68.5} & \cellcolor[HTML]{FFB366}{71.7} & \cellcolor[HTML]{FFB366}{70.1} \\

Sa2VA-InternVL3-14B 
& \cellcolor[HTML]{FFE5CC}{70.5} & \cellcolor[HTML]{FFE5CC}{74.1} & \cellcolor[HTML]{FFE5CC}{72.2} 
& \cellcolor[HTML]{FFCC99}{72.7} & \cellcolor[HTML]{FFCC99}{76.1} & \cellcolor[HTML]{FFCC99}{74.4}
& \cellcolor[HTML]{FFCC99}{74.2} & \cellcolor[HTML]{FFCC99}{77.9} & \cellcolor[HTML]{FFCC99}{76.0}
& \cellcolor[HTML]{FFE5CC}{64.5} & \cellcolor[HTML]{FFE5CC}{68.2} & \cellcolor[HTML]{FFE5CC}{66.3} \\

Sa2VA-Qwen2.5-VL-7B  
& \cellcolor[HTML]{FFCC99}{71.1} & \cellcolor[HTML]{FFCC99}{74.5} & \cellcolor[HTML]{FFCC99}{72.8}
& \cellcolor[HTML]{FFE5CC}{71.9} & \cellcolor[HTML]{FFE5CC}{74.9} & \cellcolor[HTML]{FFE5CC}{73.4}
& \cellcolor[HTML]{FFE5CC}{74.1} & \cellcolor[HTML]{FFE5CC}{77.7} & \cellcolor[HTML]{FFE5CC}{75.9}
& \cellcolor[HTML]{FFCC99}{67.3} & \cellcolor[HTML]{FFCC99}{70.8} & \cellcolor[HTML]{FFCC99}{69.1} \\

Sa2VA-Qwen3-VL-4B  
& 66.8 & 70.4 & 68.6
& 66.5 & 70.0 & 68.3
& 73.4 & 77.0 & 75.2
& 60.4 & 64.1 & 62.2 \\

\Xhline{1pt}
\end{tabularx}
} % end resizebox

\label{tab:grounding_table}
\vspace{-5mm}
\end{table*}

\subsection{Main Results}
\vspace{-1mm}
Tab.~\ref{tab:dynbench_table} and Tab.~\ref{tab:grounding_table} report the overall performance on \texttt{Dyn-Bench}, and Fig.~\ref{fig:mllms_compare} shows the radar results across the three VQA levels. Our findings are summarized as follows:

\noindent \textbf{General MLLM.}  
Proprietary models such as GPT-4o and GPT-5 maintain strong performance in spatio-temporal reasoning, particularly excelling in inter-object understanding through accurate modeling of actions and relational dynamics. In contrast, open-source models have rapidly narrowed the gap, with large-scale systems like Qwen3-VL-235B achieving comparable or even superior overall results, and smaller variants such as LLaVA-OneVision-1.5-8B and Qwen3-VL-32B delivering competitive accuracy despite reduced parameter counts. Overall, proprietary models tend to dominate relational and motion-oriented reasoning, while open-source models demonstrate more balanced generalization across object- and scene-level understanding. 

\noindent \textbf{Spatial MLLMs.}
Compared with general MLLMs, spatial models exhibit stronger performance on geometry-dependent object–scene reasoning, highlighting the value of explicit spatial priors. Within this category, VST-7B-RL attains the strongest overall performance, with SpaceR-7B and SpatialLadder-3B following closely. However, despite their strengths in static and relational spatial tasks, spatial MLLMs remain weaker than both general and region-level models on camera–object interaction and motion-centric reasoning, indicating that spatial priors alone are insufficient for modeling dynamic 4D scenes.

\noindent \textbf{Region-level MLLMs.}  
Models in this category deliver the strongest performance on object-centric spatio-temporal reasoning and dynamic object grounding, enabled by their integration of fine-grained regional cues and localized feature alignment. UniPixel-7B provides the best overall spatio-temporal reasoning within this group, while Sa2VA-based variants achieve the highest grounding accuracy across diverse dynamic settings. Relative to both general and spatial MLLMs, these models exhibit pronounced advantages in motion understanding and relational dynamics, indicating that region-level grounding supplies robust structural priors that enhance temporal coherence and support more reliable interpretation of complex dynamic scenes.

\begin{figure*}[ht]  % b = bottom
    \centering      % 或去掉改成 \raggedright 让它靠左
    \includegraphics[width=0.9\linewidth]{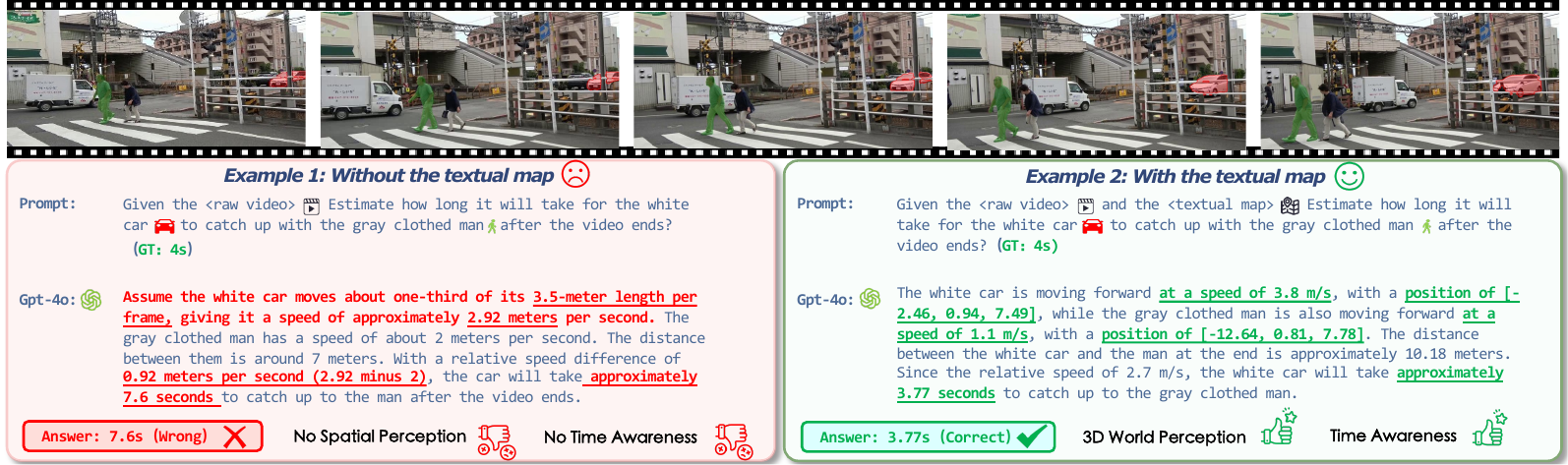}
    \caption{\textbf{Examples of how an MLLM \textit{thinks in dynamics} as revealed by self-explanations.} While the model shows strong linguistic reasoning, its 4D world modeling capability remains limited without explicit spatio-temporal guidance.}
    \label{case_compare}
    \vspace{-5mm}
\end{figure*}

\vspace{-2mm}
\section{How MLLMs \emph{Think in Dynamics} Textually}
\vspace{-1mm}
To investigate \emph{how MLLMs think in dynamics textually}, we first analyze GPT-4o’s self-explanations on failure cases from \texttt{Dyn-Bench} to examine its Chain-of-Thought (CoT) behavior in dynamic settings. We then introduce a ST-TCM as an auxiliary input to qualitatively assess its effect on reasoning. Finally, we conduct an ablation study on three key components, namely \textit{temporal semantics}, \textit{spatial geometry}, and \textit{motion dynamics}, to identify which factors most effectively enhance spatio-temporal reasoning.

\vspace{-1mm}
\subsection{Self-Explanations in Dynamic Reasoning}
\vspace{-1mm}
Self-explanation refers to an MLLM’s ability to articulate intermediate reasoning while generating responses. 
We examine GPT-4o’s self-explanations on \texttt{Dyn-Bench} and present representative success and failure cases in Fig.~\ref{case_compare} to illustrate its reasoning strengths and limitations.

\noindent \textbf{Case Studies.}  
Fig.~\ref{case_compare} compares GPT-4o's \textit{self-explanations} in a failure and a success case. In the failure case (Fig.~\ref{case_compare}, left), the model generates linguistically fluent but physically inconsistent reasoning. When estimating how a white car catches up with a gray-clothed pedestrian, it relies on visual heuristics such as apparent size change across frames rather than metric reasoning, resulting in inaccurate temporal estimation, reflecting a gap between linguistic coherence and physical grounding. In contrast, the success case (Fig.~\ref{case_compare}, right) demonstrates structured reasoning integrating motion and relational cues to estimate relative velocities and produce temporally consistent predictions. These findings suggest that explicit spatio-temporal cues enable more coherent and causally grounded reasoning about dynamic events.

\noindent \textbf{Error Analysis.}  
GPT-4o’s errors in dynamic reasoning can be broadly categorized into three fundamental types:
\ding{172} \textit{Temporal reasoning errors}, where the model fails to maintain event order or motion continuity, interpreting sequences as discrete frames rather than continuous processes;
\ding{173} \textit{Spatial grounding errors}, arising from limited geometric understanding and resulting in inaccurate distance or position estimation; and
\ding{174} \textit{Relational reasoning errors}, reflecting persistent difficulty in capturing causal or interactional dependencies between objects.
These errors indicate that GPT-4o lacks structured temporal, spatial, and relational representations, ultimately constraining its ability to reason about motion in a physically coherent manner.

\vspace{-2mm}

\subsection{Textual Guidance in Dynamics}
\vspace{-1mm}
To further investigate how ST-TCM enhances spatio-temporal reasoning and dynamic object grounding, we conduct an ablation over its three components: \textit{temporal semantics (T)}, \textit{motion dynamics (M)}, and \textit{spatial geometry (S)}. We evaluate two representative models under distinct paradigms: the general MLLM Qwen3-VL-8B for spatio-temporal reasoning and the region-level MLLM UniPixel-3B for dynamic object grounding. 
As shown in Tab.~\ref{tab:merged_ablation}, incorporating ST-TCM components consistently improves both models, though their optimal configurations differ. For Qwen3-VL-8B, motion and spatial cues (M+S) produce the largest gains, highlighting the role of object movement and geometric structure in achieving stable temporal reasoning, whereas temporal cues alone are insufficient. For UniPixel-3B, motion cues offer the primary improvement, and spatial cues mainly refine object–trajectory alignment. The M-only configuration attains the best overall performance.

\vspace{-2mm}
\begin{table*}[htpb]
\centering
\caption{\textbf{Ablation of Spatio-Temporal Textual Cognitive Map.}
Top three performers in each column are highlighted from 
\colorbox[HTML]{FFB366}{Dark} (highest) to \colorbox[HTML]{FFE5CC}{Light} (third highest). 
\textit{T}, \textit{M}, and \textit{S} denote temporal semantics, motion dynamics, and spatial geometry, respectively.}
\label{tab:merged_ablation}
\vspace{-2mm}
\setlength\tabcolsep{4pt}
\renewcommand{\arraystretch}{1.2}
\scriptsize
\resizebox{0.8\textwidth}{!}{%
\begin{tabular}{%
lcccc
@{\hspace{10pt}}|@{\hspace{10pt}} % Adjusted spacing
c@{\hspace{6pt}}c@{\hspace{8pt}}c   % Inter-Object: J / F / J&F
@{\hspace{14pt}} % Added more space between the blocks
c@{\hspace{6pt}}c@{\hspace{8pt}}c   % Object-Scene: J / F / J&F
@{\hspace{14pt}}
c@{\hspace{6pt}}c@{\hspace{8pt}}c   % Camera-Object: J / F / J&F
}
\toprule
\multirow{2}{*}{\textbf{Configuration}} &
\multirow{2}{*}{\textbf{Inter-Object}} &
\multirow{2}{*}{\textbf{Object-Scene}} &
\multirow{2}{*}{\textbf{Camera-Object}} &
\multirow{2}{*}{\textbf{Avg}} &
\multicolumn{3}{c@{\hspace{10pt}}}{\centering\textbf{Inter-Object}} &
\multicolumn{3}{c@{\hspace{14pt}}}{\centering\textbf{Object-Scene}} &
\multicolumn{3}{c}{\centering\textbf{Camera-Object}} \\
& & & & &
$\mathcal{J}$ & $\mathcal{F}$ & $\mathcal{J\&F}$ &
$\mathcal{J}$ & $\mathcal{F}$ & $\mathcal{J\&F}$ &
$\mathcal{J}$ & $\mathcal{F}$ & $\mathcal{J\&F}$ \\
\midrule
\rowcolor{navyblue!5} 
\multicolumn{5}{c@{\hspace{10pt}}|@{\hspace{10pt}}}{\textbf{\textit{Qwen3-VL-32B}}} &
\multicolumn{9}{c}{\textbf{\textit{Sa2VA-InternVL2.5-8B}}} \\
w/o TCM
& 59.0 & 76.7 & 56.2 & 62.8
& 74.8 & 76.6 & 75.2 & 78.4 & 81.7 & 80.0 & 70.3 & 73.5 & 71.9 \\
w/ T only 
& 59.3 & 76.4 & 56.3 & 62.9
& 76.6 & 79.3 & 78.0 & 79.7 & 82.9 & 81.3 & 73.0 & 76.1 & 74.6 \\
w/ M only 
& 64.3 & 77.1 & 53.5 & 63.3
& 76.8 & 79.6 & 78.2 & 79.7 & 83.0 & 81.4 &
  73.8 & 77.3 & 75.5 \\
w/ S only 
& 66.1 & \cellcolor[HTML]{FFE5CC}{78.7} & \cellcolor[HTML]{FFCC99}{60.1} & \cellcolor[HTML]{FFE5CC}{67.2}
& 76.9 & 79.7 & 78.3 &
  79.9 & 83.2 & 81.5 &
  74.8 & 78.5 & 76.4 \\
w/ T + M 
& 63.8 & 76.7 & 54.0 & 63.3
& \cellcolor[HTML]{FFE5CC}{77.0} & \cellcolor[HTML]{FFE5CC}{79.8} & \cellcolor[HTML]{FFE5CC}{78.4} & 79.8 & 83.1 & 81.4 & 73.8 & 77.3 & 75.5 \\
w/ T + S 
& \cellcolor[HTML]{FFE5CC}{67.0} & 78.5 & \cellcolor[HTML]{FFE5CC}{59.6} & 67.1
& 76.9 & 79.7 & 78.3 & \cellcolor[HTML]{FFE5CC}{80.0} & \cellcolor[HTML]{FFE5CC}{83.3} & \cellcolor[HTML]{FFCC99}{81.8} & \cellcolor[HTML]{FFE5CC}{74.9} & \cellcolor[HTML]{FFE5CC}{78.6} & \cellcolor[HTML]{FFE5CC}{76.7} \\
w/ M + S 
& \cellcolor[HTML]{FFCC99}{68.4} & \cellcolor[HTML]{FFCC99}{78.8} & 59.4 & \cellcolor[HTML]{FFCC99}{67.5}
& \cellcolor[HTML]{FFCC99}{77.1} & \cellcolor[HTML]{FFCC99}{79.9} & \cellcolor[HTML]{FFCC99}{78.5} & \cellcolor[HTML]{FFCC99}{80.1} & \cellcolor[HTML]{FFCC99}{83.5} & \cellcolor[HTML]{FFE5CC}{81.6} & \cellcolor[HTML]{FFCC99}{75.3} & \cellcolor[HTML]{FFCC99}{78.9} & \cellcolor[HTML]{FFCC99}{77.1} \\
w/ T + M + S 
& \cellcolor[HTML]{FFB366}{69.2} & \cellcolor[HTML]{FFB366}{79.1} & \cellcolor[HTML]{FFB366}{60.5} & \cellcolor[HTML]{FFB366}{68.3}
& \cellcolor[HTML]{FFB366}{77.3} & \cellcolor[HTML]{FFB366}{80.2} & \cellcolor[HTML]{FFB366}{78.8} &
  \cellcolor[HTML]{FFB366}{80.2} & \cellcolor[HTML]{FFB366}{83.6} & \cellcolor[HTML]{FFB366}{81.9} &
  \cellcolor[HTML]{FFB366}{75.4} & \cellcolor[HTML]{FFB366}{79.1} & \cellcolor[HTML]{FFB366}{77.3} \\
\bottomrule
\end{tabular}%
}
\vspace{-5mm}
\end{table*}

\section{How MLLMs \emph{Think in Dynamics} Visually}

\begin{figure}[h]  % b = bottom
    \centering      % 或去掉改成 \raggedright 让它靠左
    \includegraphics[width=1.0\linewidth]{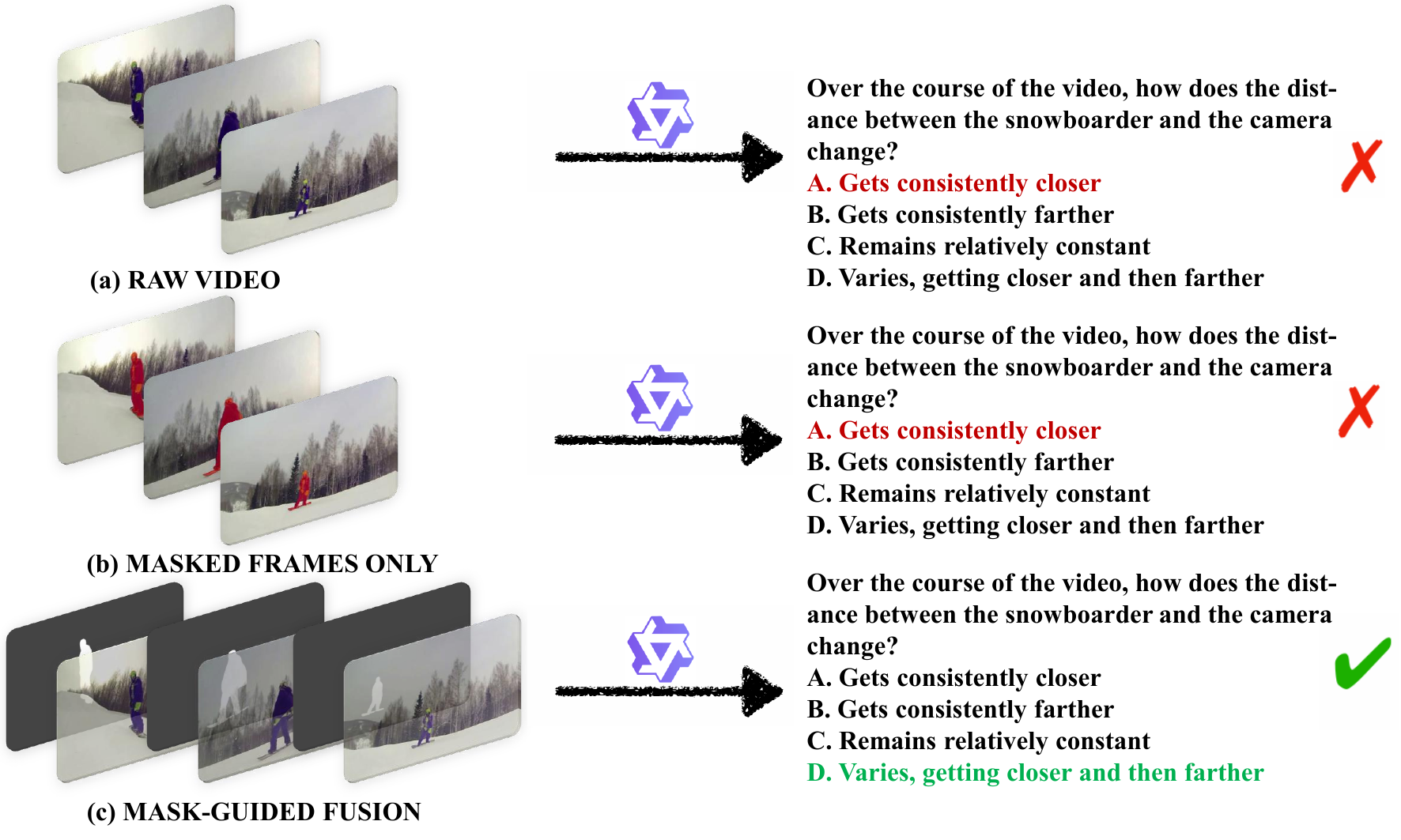}
    \caption{\textbf{Mask-Guided Input Comparison.}}
    \label{fig:mask_compare}
    \vspace{-3mm}
\end{figure}

\vspace{-2mm}
To examine how MLLMs \textit{think in dynamics} visually, we conduct qualitative and quantitative analyses to study how explicit visual guidance affects motion understanding.
As shown in Fig.~\ref{fig:mask_compare}, we design two input strategies to guide model attention toward dynamic regions.
\ding{172} \textit{Masked Frames Only} overlays object segmentation masks on each frame, directing attention to moving entities while maintaining temporal continuity.
\ding{173} \textit{Mask-Guided Fusion} combines raw frames with their corresponding masks, integrating complementary cues from appearance and motion.
These strategies explicitly ground visual perception in motion-centric regions, enhancing spatio-temporal alignment and relational reasoning.
We evaluate Qwen3-VL-8B under these configurations, using the \textit{Raw Video} setting as baseline.
\vspace{-1mm}

The results in Tab.~\ref{tab:mask_ablation} show that mask-guided visual grounding enhances the model’s ability to capture dynamic object interactions and maintain temporal consistency. \textit{Masked Frames Only} offers only minor gains, suggesting limited value from isolated localization cues. In contrast, the \textit{Mask-Guided Fusion} setting improves all categories, with the largest gains in \textit{Inter-Object} and \textit{Camera-Object} reasoning, which demand fine-grained motion and relational understanding. These results show that integrating appearance and motion cues strengthens object grounding and yields more coherent spatio-temporal reasoning.

\begin{table}[htpb]
\centering

\caption{\textbf{Quantitative Comparison of Mask-Guided Inputs.}}
\vspace{-2mm}
\setlength\tabcolsep{3pt} % tighter column spacing
\renewcommand{\arraystretch}{0.95} % slightly tighter row spacing
\scriptsize % smaller font for single-column fit
% \vspace{1mm}
\begin{tabular}{lcccc}
\toprule
\textbf{Configuration} & \textbf{Inter-Object} & \textbf{Object-Scene} & \textbf{Camera-Object} & \textbf{Avg} \\
\midrule
\textsc{Raw Video} & 38.9 & 74.5 & 55.6 & 53.8 \\
\textsc{Masked Frames Only} & 39.4 & 74.3 & 54.9 & 53.8 \\
\textsc{Mask-Guided Fusion} & \textbf{41.8} & \textbf{77.0} & \textbf{60.0} & \textbf{57.1} \\
\bottomrule
\end{tabular}
\vspace{-5mm}
\label{tab:mask_ablation}
\end{table}
\vspace{1mm}

\vspace{-2mm}

\section{Related Work}
\label{sec:related_work}
\vspace{-1mm}
\noindent \textbf{Spatio-Temporal Understanding with MLLMs.}
Recent MLLMs~\cite{Gemini2.5, gpt-5, Gpt-4o, Qwen3-Omni, InternVL3.5, LLaVA-OneVision-1.5, Qwen2.5-VL, GLM-4.5V} have advanced from static visual perception to general spatio-temporal reasoning. Early video-focused models~\cite{Video-llava, Videochat} introduced temporal encodings for short-term scene understanding and basic motion interpretation. General MLLMs~\cite{Qwen3-Omni, InternVL3.5, Gemini2.5, gpt-5} further improve temporal awareness through adaptive attention and resolution routing, enabling context-sensitive reasoning over continuous frames. Spatial MLLMs~\cite{spacer, vst, spatial-ssrl, spatailvlm, spatialmllm, spatialcot, scener1, 3dr1, sr3d, vlm3r} leverage geometric priors and 3D representations for spatial reasoning, yet they remain largely restricted to static or indoor environments without modeling dynamic scene evolution. Building on this, recent 4D reasoning frameworks~\cite{UnderstandingDynamicEgoCentric4D, LLaVA-4D, Uni4D-LLM} further extend multimodal understanding into complex, dynamic real-world settings by leveraging RGB-D cues and cross-attentional integration across temporal and geometric dimensions.

\vspace{0.2em}
\noindent \textbf{Region-level MLLMs.}
Recent progress has advanced MLLMs toward region-level perception and reasoning, enabling finer-grained, context-aware understanding. These models integrate segmentation, referring, and reasoning modules, connecting localized regions with high-level semantics. Sa2VA~\cite{Sa2va} improves token--region alignment via cross-modal correspondence, PAM~\cite{PerceiveAnything} refines multi-scale features, GAR~\cite{GraspAnyRegion} models inter-prompt dependencies, and UniPixel~\cite{Unipixel} unifies region grounding and mask prediction end-to-end. PixelRefer~\cite{pixelrefer} and VideoRefer~\cite{Videorefer} enhance region-level grounding through efficient architectures and supervision, while DAM~\cite{Describeanything} strengthens localized captioning with focal prompting and region-sensitive encoders.
Despite these advances, stable spatial grounding and temporal coherence in dynamic scenes remain challenging for region-level MLLMs, especially under rapid motion, occlusion, or complex interactions.

\vspace{0.2em}
\noindent \textbf{Benchmarks for Spatio-Temporal Intelligence.}
Most existing spatio-temporal benchmarks~\cite{DSI-Bench, VSI-bench, ODI-Bench, Ost-bench, Vlm4d, Sti-bench} emphasize scene-level or observer-centric reasoning, providing coarse evaluations of model understanding in dynamic environments.  
STI-Bench~\cite{Sti-bench} focuses on quantitative motion reasoning in dynamic indoor and outdoor scenes;  
OST-Bench~\cite{Ost-bench} explores agent-centric temporal reasoning through continuous observation and interaction;  
VLM4D~\cite{Vlm4d} investigates 4D spatial and temporal awareness in real and synthetic videos;  
and DSI-Bench~\cite{DSI-Bench} analyzes spatial consistency under coupled camera and object motion.  
However, these benchmarks lack systematic evaluation of dynamic object understanding, particularly in modeling fine-grained motion and temporal continuity.
\vspace{-2mm}
\section{Discussion and Future Work}
\vspace{-1mm}
We study how MLLMs \textit{think in dynamics} by introducing \texttt{Dyn-Bench}, a benchmark evaluating object-level and scene-level spatio-temporal reasoning and grounding across model categories: general, spatial, and region-level MLLMs. Through this dual textual--visual assessment, we systematically examine how models perceive, track, and interpret dynamic content in the physical 4D world, including capturing motion patterns, maintaining temporal consistency, and modeling multi-entity interactions.
Our experiments show that the Spatio-Temporal Textual Cognitive Map substantially enhances temporal coherence and relational reasoning through structured linguistic abstraction of dynamic events, while mask-guided visual grounding strengthens motion perception, improves fine-grained object continuity, and mitigates temporal drift. These findings suggest that reliable dynamic understanding in MLLMs emerges from coupling high-level temporal semantics with localized region-level grounding.
Future spatio-temporal MLLMs should tightly integrate dynamic-object perception and temporal reasoning, motivating unified architectures that jointly model motion dynamics, relational structure, and higher-level temporal cognition for coherent, physically grounded reasoning in complex, evolving environments.

\newpage
{
    \small
    \bibliographystyle{ieeenat_fullname}
    \bibliography{ref}
}

\clearpage
\setcounter{page}{1}
\maketitlesupplementary

% This is the supplementary material for the paper:  
% \textit{``Thinking in Dynamics: How Multimodal Large Language Models Perceive, Track, and Reason Dynamics in the Physical 4D World.''}  
% We provide additional implementation details, visualizations, and prompt templates as follows:

% \begin{itemize}
%     \item \textbf{A. More Implementation Details.}
%     \begin{itemize}
%         \item A.1 Data Filter Strategy.
%         \item A.2 Spatio-Temporal Textual Cognitive Map (ST-TCM).
%     \end{itemize}

%     \item \textbf{B. More Visual Results.}
%     \begin{itemize}
%         \item B.1 Qualitative Examples of \texttt{Dyn-Bench}.
%         \item B.2 Failure Cases on \texttt{Dyn-Bench}.
%     \end{itemize}

%     \item \textbf{C. Prompt Templates for Data Generation.}
% \end{itemize}

% Supplementary introduction with working refs
This is the supplementary material for the paper:  
\textit{``Thinking in Dynamics: How Multimodal Large Language Models Perceive, Track, and Reason Dynamics in the Physical 4D World.''}  
It provides additional implementation details, visualizations, and prompt templates to complement the main paper. The contents are organized as follows:

\begin{itemize}[leftmargin=1.5em]
    \item In \underline{{\color{blue}Section{\color{red}~\ref{sec:implementation}}}}, we detail the implementation of our framework, including the data filter strategy for dynamic video curation and the construction of the \textit{Spatio-Temporal Textual Cognitive Map (ST-TCM)} for object-centric reasoning.

    \item In \underline{{\color{blue}Section{\color{red}~\ref{sec:visual}}}}, we present qualitative visualizations from \texttt{Dyn-Bench}, including examples across hierarchical levels, representative failure cases, and comparisons highlighting the effects of \textit{ST-TCM} and \textit{Mask-Guided Input}.

    \item In \underline{{\color{blue}Section{\color{red}~\ref{sec:prompt}}}}, we describe six prompt templates used in the QA generation stage for creating Visual Question Answering (VQA) and grounding data across dynamic perception and reasoning tasks.
\end{itemize}

% ------------------------
% Section A Configuration
\renewcommand{\thesection}{A}
\renewcommand{\thesubsection}{A.\arabic{subsection}}

\section{More Implementation Details}
\addcontentsline{toc}{section}{A. More Implementation Details}
\label{sec:implementation}

\subsection{Data Filter Strategy}
\label{sec:implementation_filter}

The goal of our video data filtering process is to identify videos that exhibit rich spatio-temporal dynamics, including object and camera motion, inter-object interactions, human activities, and sequential manipulations, while excluding static, artificial, or heavily edited content such as landscapes, cartoons, or composited clips.

To ensure that the curated dataset effectively supports \textit{spatio-temporal reasoning} and \textit{dynamic object grounding}, we define three key selection criteria:

\begin{enumerate}[label=\textbf{S\arabic*.}]
    \item \textbf{High-quality video content.}  
    Videos should maintain sufficient visual fidelity in terms of resolution, frame rate, and perspective stability, without severe distortion or over-processing artifacts.Shaky frames or inconsistent frame rates should also be excluded.
    
    \item \textbf{Feasibility for motion and geometry reasoning.}  
    Videos should be suitable for reliable geometric estimation and motion reasoning. We exclude clips with extreme zooming, abrupt shot transitions, or ambiguous reference frames, such as those captured from moving vehicles. Videos that lack stable visual correspondences, including those with heavily blurred or fully occluded backgrounds, are also removed from the dataset.

    \item \textbf{Dynamic camera and scene motion.}  
    Videos should contain non-static cameras or dynamically evolving scenes that allow non-trivial spatio-temporal reasoning. Such videos often capture human--object or inter-object interactions, providing diverse motion for downstream reasoning and grounding tasks.
\end{enumerate}

To meet these requirements, we design a multi-stage filtering pipeline integrating motion, geometry, dynamic coverage, and multimodal semantics. Each stage progressively refines dataset quality to ensure geometric consistency and temporal richness.

\begin{figure*}[t]
    \centering
    \includegraphics[width=0.9\textwidth]{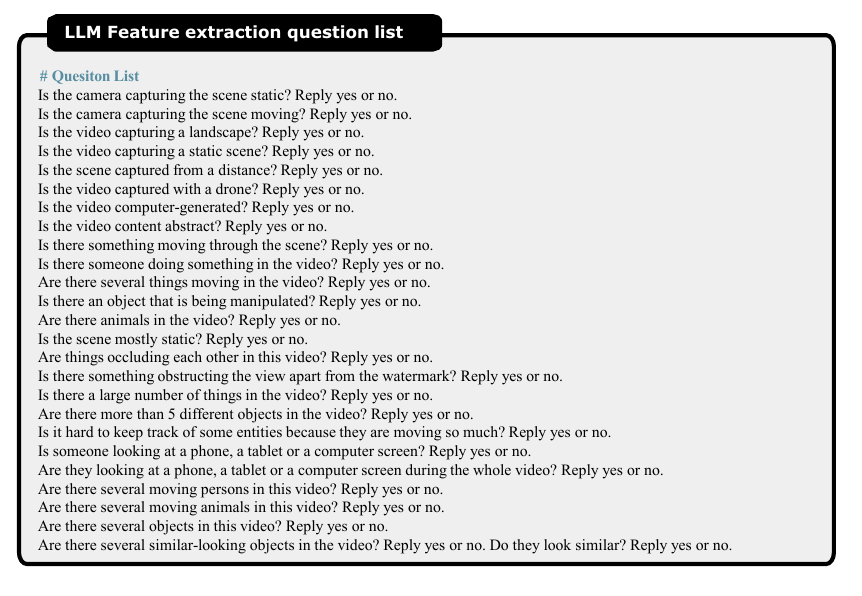}
    \vspace{-2mm}
    \caption{LLM Feature Extraction: the model answers 26 structured diagnostic questions to extract semantic and motion cues.}
    \label{fig:26q}
\end{figure*}

\begin{enumerate}[label=\textbf{S\arabic*.},resume]
    \item \textbf{Compute motion features.}  
    For each video, we extract low-level motion and quality statistics, including blur degree, frame rate (fps), number of I-frames, motion vector magnitude, and motion vector variance, using OpenCV. These features characterize the global motion intensity and temporal smoothness of the video, serving as primary indicators of scene dynamics.

    \item \textbf{Obtain geometric features.}  
    Using VGGT~\cite{vggt} and UniDepth-V2~\cite{unidepthv2}, we estimate per-frame camera intrinsics, extrinsics, and depth maps to evaluate geometric stability across time. We further assess depth continuity, focal-length stability, and camera motion smoothness to ensure temporally coherent geometry. For 2D datasets without ground-truth geometry, these models also reconstruct camera pose and depth to provide metric-scale estimation.
    
    \item \textbf{Dynamic object coverage estimation.}  
    We estimate the overall degree of scene dynamics by measuring the proportion and spatial distribution of moving objects across frames, which captures both temporal motion diversity and spatial activity density. For 4D datasets lacking instance segmentation masks, Qwen2.5-VL~\cite{Qwen2.5-VL} and Sa2VA~\cite{Sa2va} are employed to infer or refine missing masks, ensuring complete and consistent instance-level motion representation throughout the sequence.

    \item \textbf{Multimodal LLM feature extraction.}  
    A multimodal large language model (MLLM), such as Qwen2.5-VL~\cite{Qwen2.5-VL}, is used to extract semantic and motion-aware features. The model answers 26 structured diagnostic questions (see Fig.~\ref{fig:26q}), capturing motion patterns, interactions, and causal relations among entities.

    \item \textbf{Video quality scoring via Random Forest.}  
    We aggregate all extracted features, comprising 31 dimensions from the preceding modules, and use them as input to a random forest regressor predicting a continuous \textit{video dynamism score} between 0 and 5.  
    A total of 3,000 videos are manually annotated for training, where 0 represents static scenes and 5 corresponds to highly dynamic, temporally complex motion.  
    The random forest jointly leverages motion, geometric, and semantic cues to estimate video-level dynamism.

    \item \textbf{VLM-assisted refinement.}  
    After filtering by predicted scores, we employ a Vision-Language Model (VLM) to further assess semantic coherence, realism, and motion validity. Only videos passing both quantitative and semantic evaluations are retained as the final high-quality dynamic dataset.
\end{enumerate}

This hierarchical filtering pipeline ensures that the resulting dataset maintains visual fidelity, geometric stability, and motion diversity, providing robust supervision for \textit{spatio-temporal reasoning} and \textit{dynamic object grounding} in the proposed \texttt{Dyn-Bench}.

\subsection{Spatio-Temporal Textual Cognitive Map (ST-TCM)}
\label{sec:implementation_sttcm}

To model fine-grained object motions and interactions in dynamic scenes, we construct a \textit{Spatio-Temporal Textual Cognitive Map (ST-TCM)} for each filtered video.  
The ST-TCM provides a unified representation that bridges geometric perception, temporal reasoning, and linguistic abstraction.  
It encodes 3D object trajectories, camera motion, and inter-object relations into structured textual descriptions, enabling object-centric spatio-temporal understanding for multimodal reasoning.  
Inspired by recent structured spatio-temporal representations and cognitive mapping approaches~\cite{VSI-bench,spatialmap,abstract3d}, we design the ST-TCM to capture dynamic interactions through both geometric grounding and textual abstraction, facilitating interpretable reasoning over temporal evolution in complex scenes.

\begin{enumerate}[label=\textbf{S\arabic*.}]
    \item \textbf{Geometric and motion reconstruction.}  
    Each video is processed at 6 FPS with synchronized RGB-D frames and instance segmentation masks to ensure balanced temporal sampling and spatial fidelity.  
    To recover metric-scale geometry and accurate camera poses, we employ VIPE~\cite{vipe}, which jointly estimates per-frame depth $\hat{D}_t$ and camera transformation $\hat{T}_t = [\hat{R}_t | \hat{\mathbf{t}}_t]$ from monocular sequences under scale-aligned supervision.  
    The 3D position of each object $o_i$ is computed by projecting its instance centroid $\tilde{\mathbf{u}}_t^i$ into world coordinates as $\mathbf{p}_t^i = \hat{T}_t^{-1} K^{-1} \tilde{\mathbf{u}}_t^i \hat{D}_t(\tilde{\mathbf{u}}_t^i)$, where $K$ denotes the camera intrinsic matrix.  
    Temporal differencing yields object velocity $\mathbf{v}_t^i = (\mathbf{p}_t^i - \mathbf{p}_{t-1}^i)/\Delta t$ and acceleration $\mathbf{a}_t^i = (\mathbf{v}_t^i - \mathbf{v}_{t-1}^i)/\Delta t$.  
    To mitigate frame-level depth noise and maintain temporal coherence, an exponential moving average is applied for smoothing.  
    This stage yields geometrically consistent and temporally stable trajectories, serving as the foundation for subsequent motion reasoning and relational inference.

    \item \textbf{Dynamic relation and spatial reasoning.}  
    Building on reconstructed trajectories, we model both inter-object and camera–object relations according to spatial proximity, relative motion, and temporal continuity.  
    For each object pair $(o_i, o_j)$, we compute their 3D Euclidean distance and evaluate a relative motion rate derived from their velocity vectors and positional difference, which distinguishes \textit{approaching}, \textit{receding}, and \textit{parallel} motion behaviors.  
    Additionally, each object’s spatial configuration relative to the camera is characterized by its azimuth and elevation angles, derived from the relative 3D coordinates, providing directional cues for qualitative reasoning such as \textit{front, left, right,} or \textit{back}.  
    This unified formulation captures both egocentric and allocentric spatial layouts, enabling consistent modeling of spatial dependencies across frames.  
    \item \textbf{Textual cognitive mapping.}  
    All geometric, kinematic, and relational attributes are systematically transformed into structured textual form through a rule-based cognitive mapping module.  
    This process integrates multi-level cues, including object geometry (size and position), motion states (direction, velocity, acceleration), and relational context (distance, orientation, and interaction type), into concise, interpretable frame-level textual representations.  
    These frame-wise textual embeddings are temporally aligned and sequentially aggregated, forming a coherent spatio-temporal narrative that maintains motion continuity and object identity throughout the sequence.  
    The resulting \textit{Spatio-Temporal Textual Cognitive Map (ST-TCM)} provides a unified symbolic–textual representation of scene dynamics, which is subsequently utilized as structured input to \textit{Qwen3-VL-235B}~\cite{Qwen3-Omni} for object-centric reasoning, dynamic scene understanding, and spatio-temporal question answering.
\end{enumerate}

% ------------------------
\renewcommand{\thesection}{B}
\renewcommand{\thesubsection}{B.\arabic{subsection}}

\section{More Visual Results}
\addcontentsline{toc}{section}{More Visual Results}
\label{sec:visual}

In this section, we present additional qualitative results to further illustrate the effectiveness of our framework.  
Comprehensive visualizations are provided across the three hierarchical levels and nine sub-tasks defined in \texttt{Dyn-Bench}, illustrating the diversity of task settings and dynamic scene types encompassed by the benchmark.
We also include representative failure cases to analyze common challenges in dynamic scene understanding.  
Furthermore, comparative visual examples are shown to evaluate the impact of the proposed \textit{Spatio-Temporal Textual Cognitive Map} and \textit{Mask-Guided Input} on different multimodal large language models, highlighting their respective strengths and limitations.

\subsection{Qualitative Examples on \texttt{Dyn-Bench}}
\label{sec:visual_examples}
\begin{figure*}[t]
    \centering
    \includegraphics[width=0.9\textwidth]{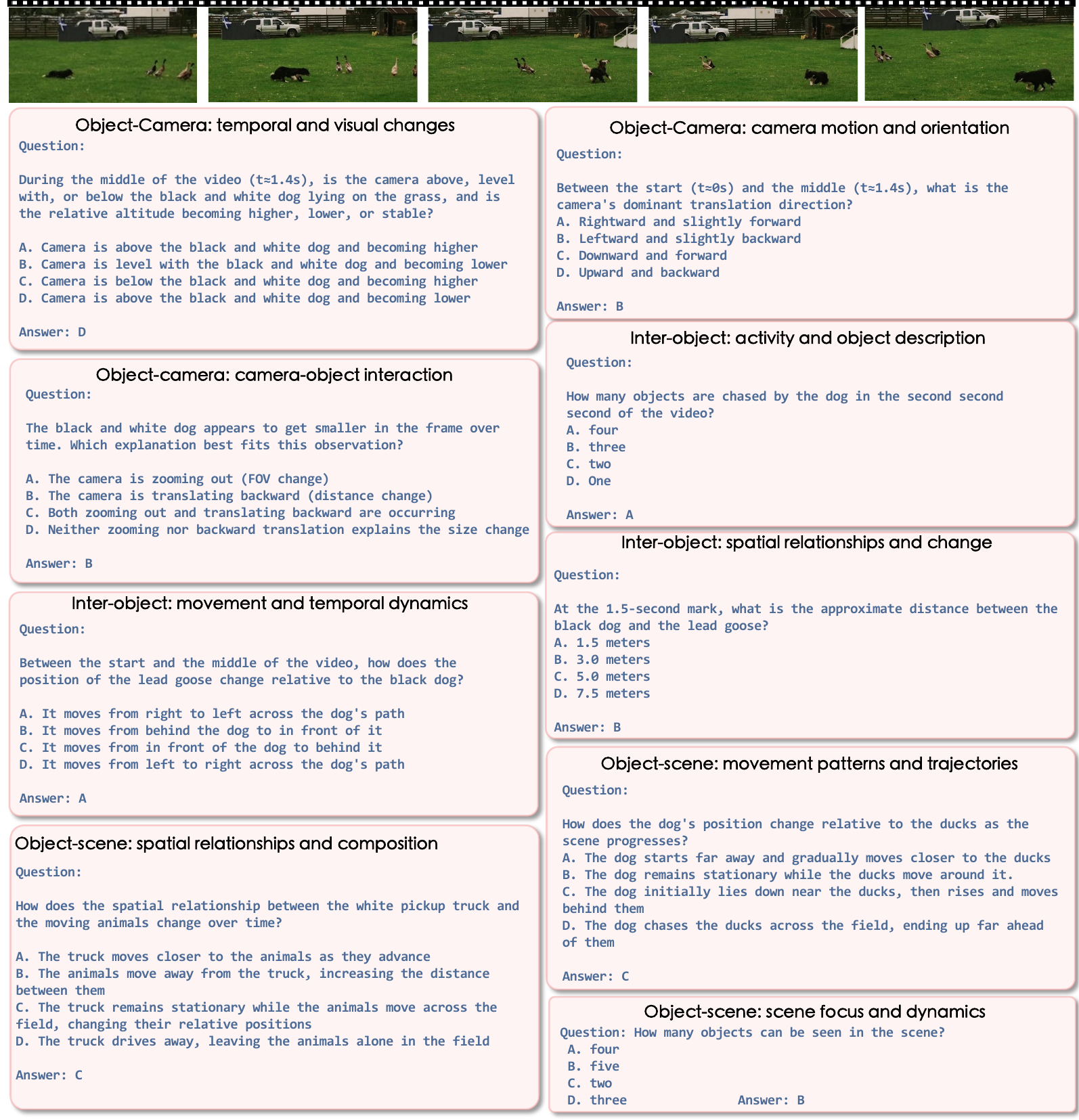}
    \vspace{-2mm}
    \caption{Qualitative examples of the nine representative dynamic understanding tasks across the three hierarchical levels of \texttt{Dyn-Bench}.}
    \label{fig:9example}
\end{figure*}

We provide qualitative examples across the three hierarchical levels of \texttt{Dyn-Bench}, including \textit{Dynamic Inter-Object Perception}, \textit{Dynamic Object-Scene Tracking}, and \textit{Dynamic Camera-Object Reasoning} (see Fig.~\ref{fig:9example}). 
Each level captures distinct dimensions of dynamic scene understanding: the first focuses on relational reasoning between dynamic objects, the second emphasizes object-scene interactions and compositional motion, and the third involves reasoning under varying camera motion and viewpoint changes. 
Within these levels, multiple sub-tasks are defined, covering nine representative types of dynamic understanding scenarios such as temporal changes, spatial alignment, activity recognition, and cross-object motion dynamics. 
Together, these examples highlight the diversity and complexity of task settings in \texttt{Dyn-Bench}, illustrating how the benchmark provides a comprehensive evaluation framework for assessing visual reasoning under dynamic and multi-object conditions. 
They also provide qualitative insights into how models interpret temporal cues, preserve semantic coherence, and adapt to scene variations in realistic videos.

\subsection{Failure Cases on \texttt{Dyn-Bench}}
\label{sec:visual_failure}
\begin{figure*}[t]
    \centering
    \includegraphics[width=0.9\textwidth]{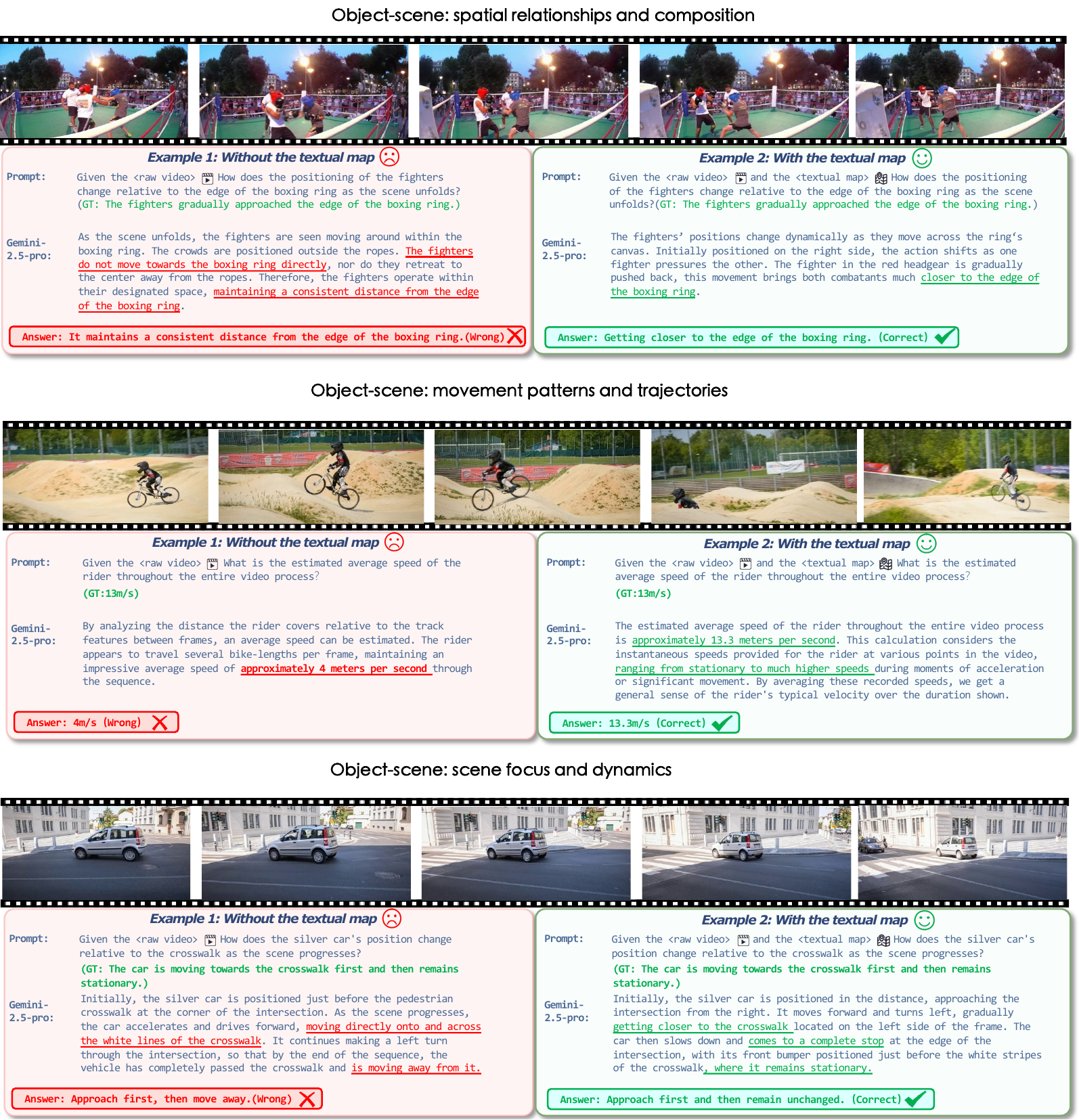}
    \vspace{-2mm}
    \caption{Representative failure cases of \textit{Gemini-2.5 Pro} on \texttt{Dyn-Bench}, showing qualitative comparisons incorporating \textit{ST-TCM}.}
    \label{fig:faliurecase_map1}
\end{figure*}

\begin{figure*}[t]
    \centering
    \includegraphics[width=0.9\textwidth]{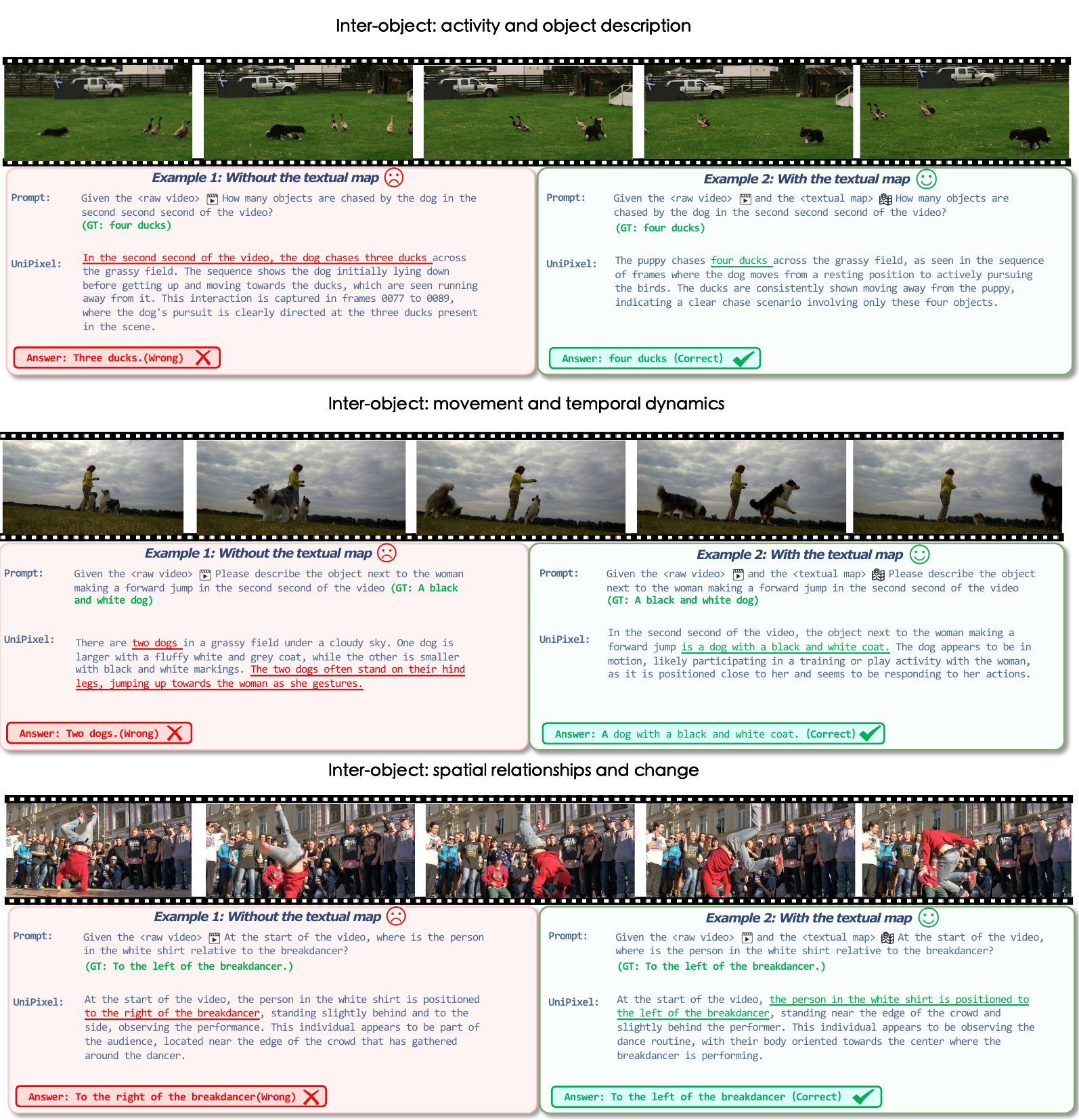}
    \vspace{-2mm}
 \caption{Representative failure cases of \textit{UniPixel-7B} on \texttt{Dyn-Bench}, showing qualitative comparisons incorporating \textit{ST-TCM}.}
    \label{fig:faliurecase_map2}
\end{figure*}

\begin{figure*}[t]
    \centering
    \includegraphics[width=0.9\textwidth]{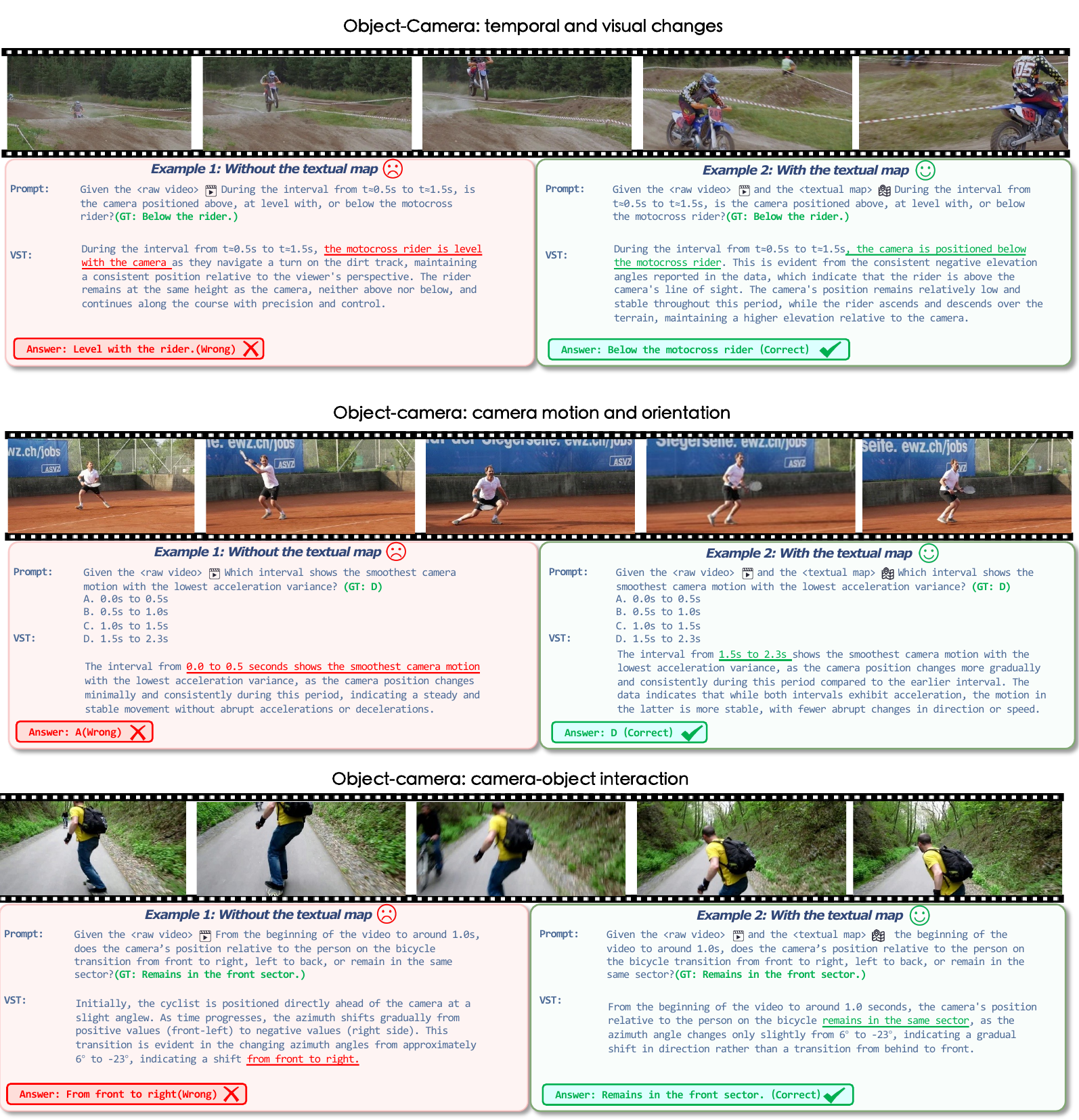}
    \vspace{-2mm}
 \caption{Representative failure cases of \textit{VST-7B-RL} on \texttt{Dyn-Bench}, showing qualitative comparisons incorporating \textit{ST-TCM}.}
    \label{fig:faliurecase_map3}
\end{figure*}

\begin{figure*}[t]
    \centering
    \includegraphics[width=0.9\textwidth]{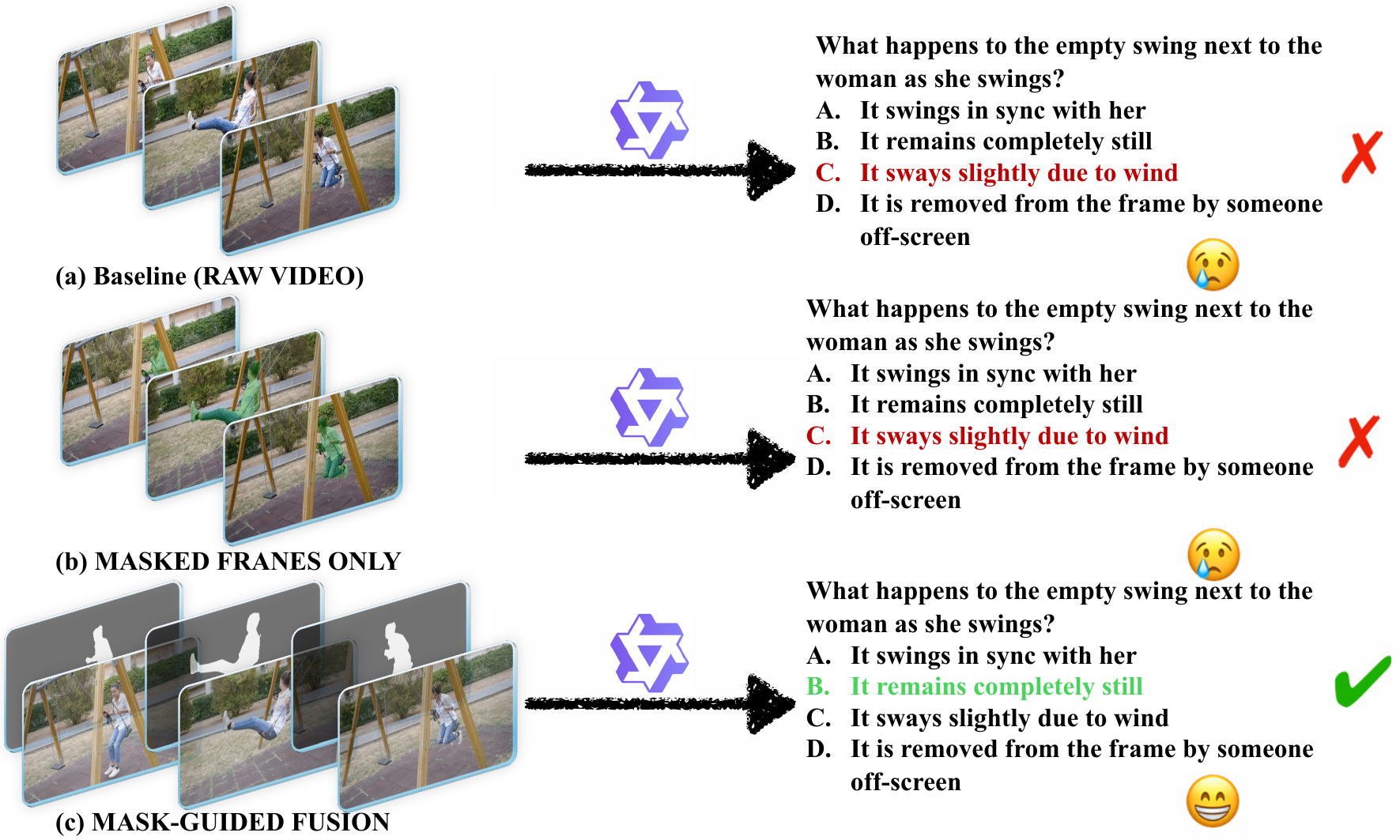}
    \vspace{-2mm}
    \caption{More visual comparison results on \textit{Mask-Guided Input}.}
    \label{fig:faliurecase_mask1}
\end{figure*}

\begin{figure*}[t]
    \centering
    \includegraphics[width=0.9\textwidth]{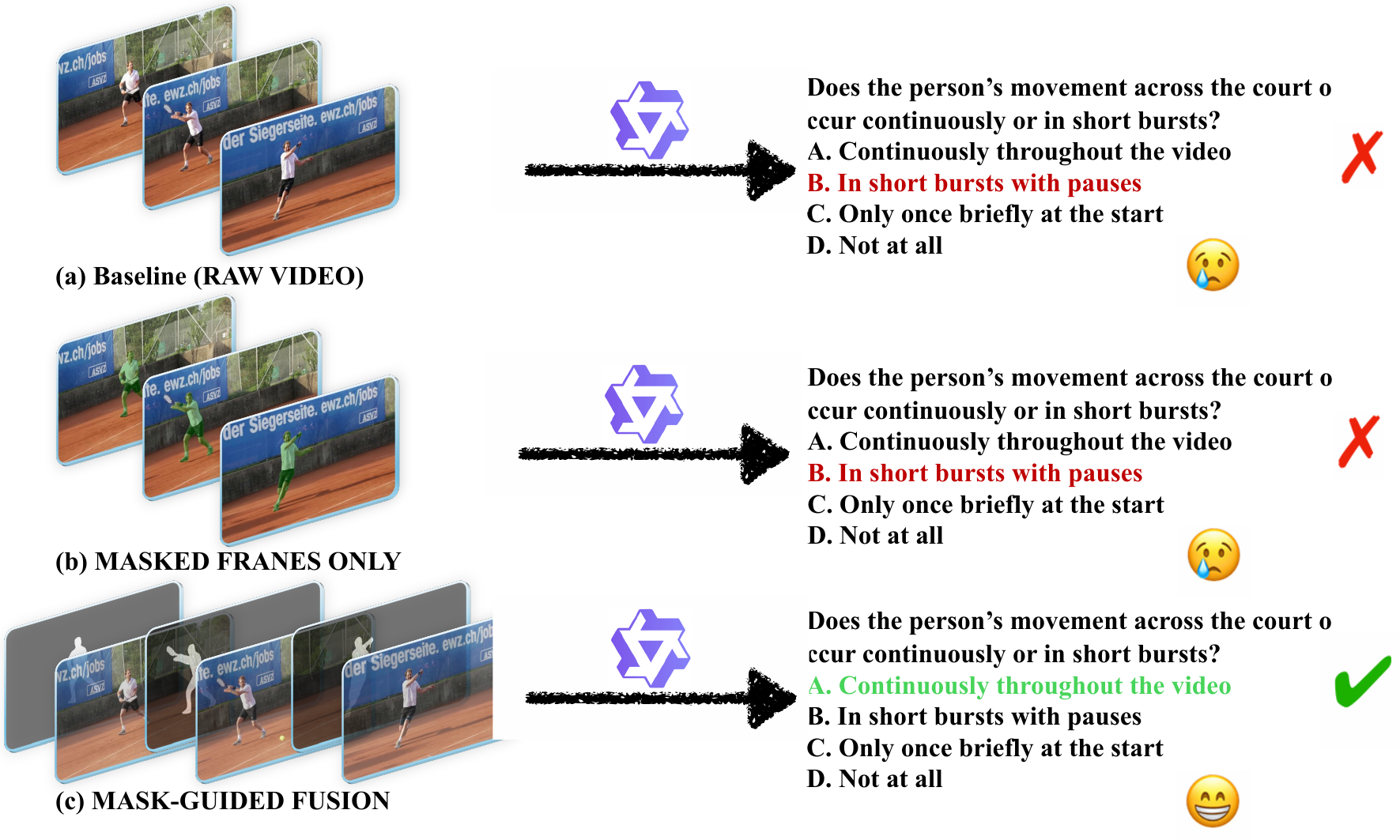}
    \vspace{-2mm}
    \caption{More visual comparison results on \textit{Mask-Guided Input}.}
    \label{fig:faliurecase_mask2}
\end{figure*}

\begin{figure*}[t]
    \centering
    \includegraphics[width=0.9\textwidth]{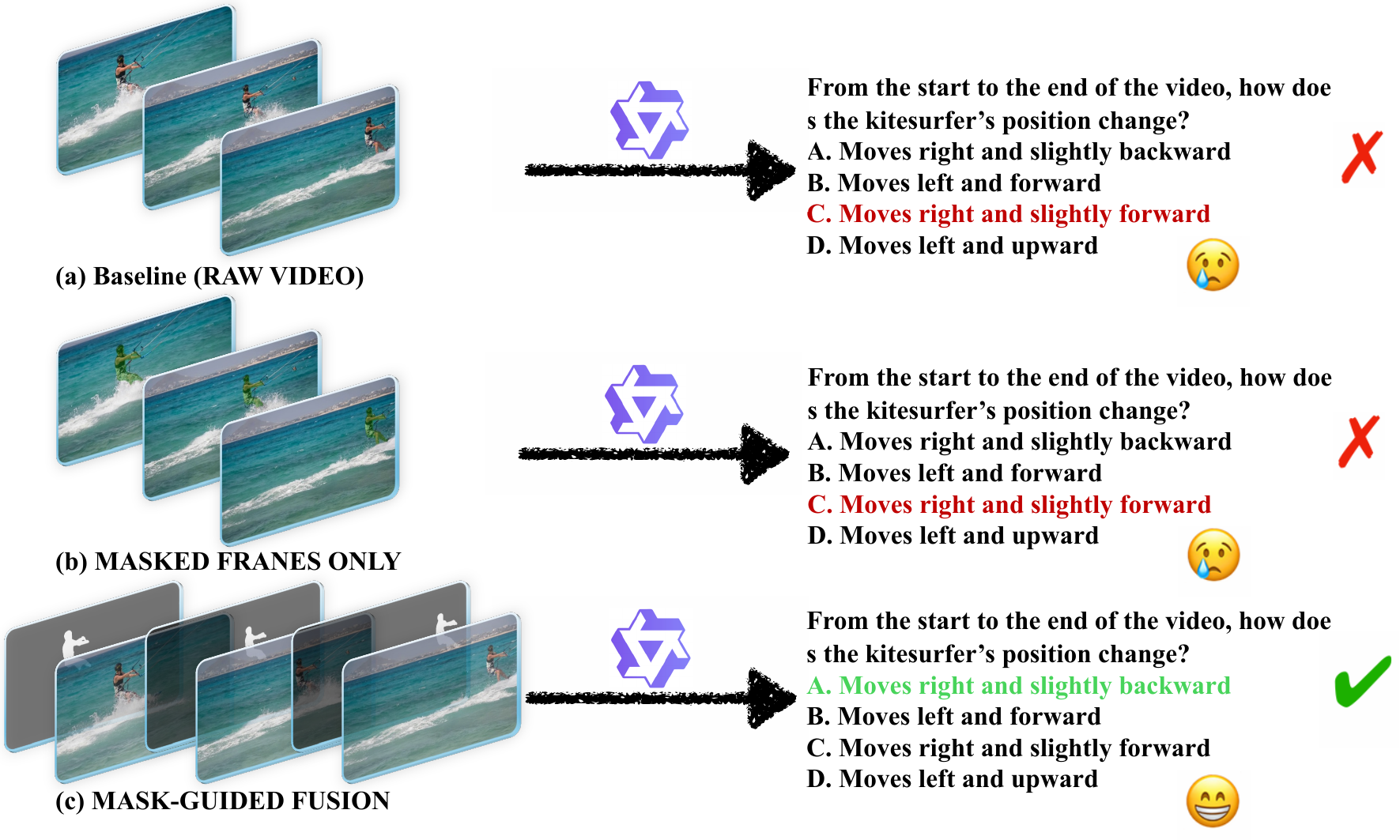}
    \vspace{-2mm}
    \caption{More visual comparison results on \textit{Mask-Guided Input}.}
    \label{fig:faliurecase_mask3}
\end{figure*}

We present failure cases from three representative categories of multimodal large language models (MLLMs), namely \textit{Gemini-2.5 Pro}~\cite{Gemini2.5} as a general MLLM, \textit{UniPixel-7B}~\cite{Unipixel} as a region-level MLLM, and \textit{VST-7B-RL}~\cite{vst} as a spatial MLLM, evaluated across the nine sub-tasks of \texttt{Dyn-Bench}. 
These cases reveal the typical reasoning and perception errors encountered when handling complex dynamic scenes, such as inaccurate temporal correlation, misalignment in spatial grounding, or failure to infer cross-object motion relationships. 
In addition, we provide qualitative comparisons using \textit{ST-TCM} and \textit{Mask-Guided Inputs} to demonstrate their effects in enhancing temporal consistency, spatial focus, and reasoning robustness across different MLLM architectures. 
Representative visual examples are shown in Fig.~\ref{fig:faliurecase_map1}–\ref{fig:faliurecase_mask3}, illustrating the limitations of existing MLLMs and the benefits of the proposed enhancements for dynamic reasoning.

\renewcommand{\thesection}{C}
\section{Prompt Templates for Data Generation}
\addcontentsline{toc}{section}{C. Prompt Templates for Data Generation}
\label{sec:prompt}

\begin{figure*}[t]
    \centering
    \includegraphics[width=0.9\textwidth]{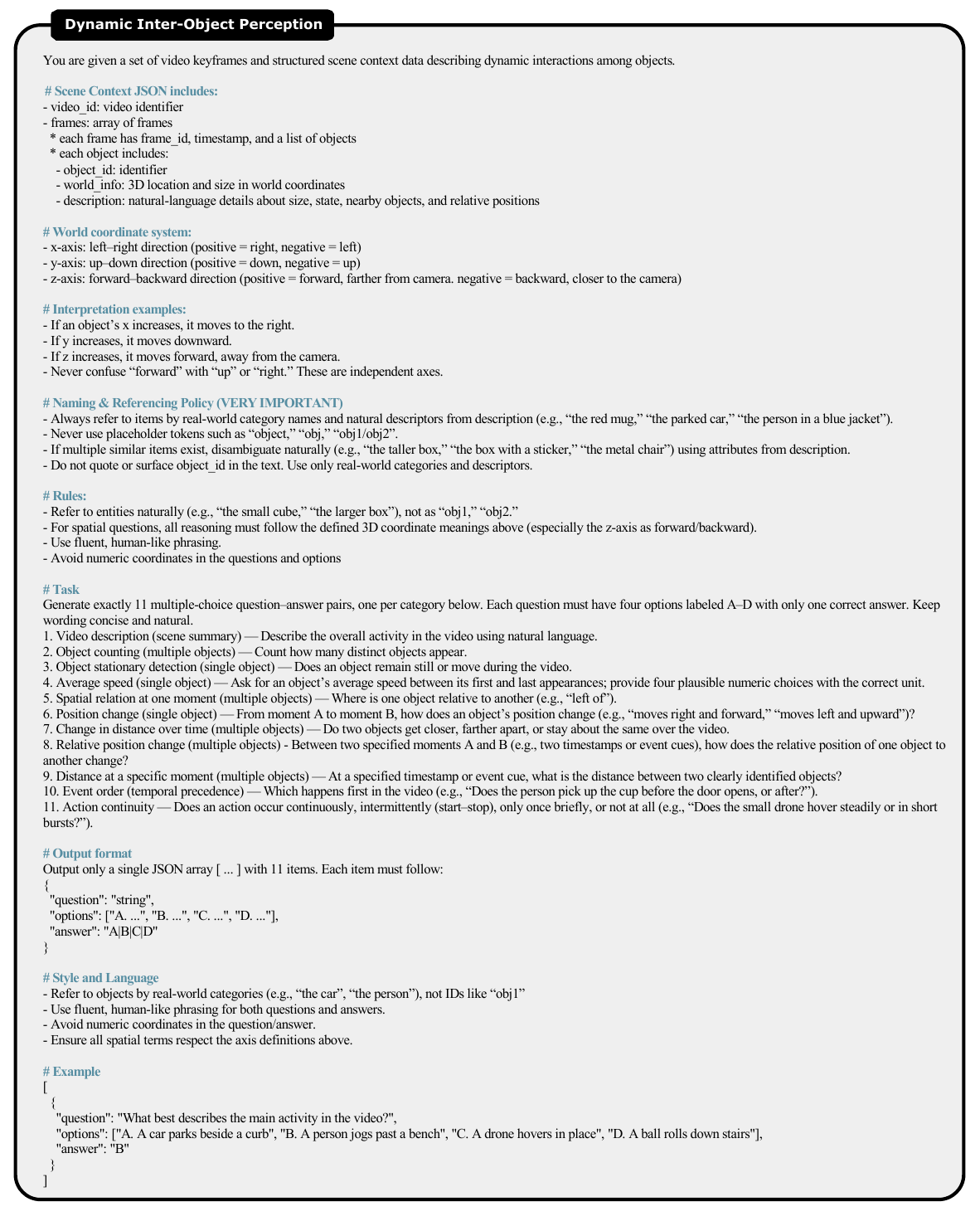}
    \vspace{-2mm}
    \caption{Prompt template for \textit{Dynamic Inter-Object Perception} in \texttt{Dyn-Bench}. }
    \label{fig:prompt1}
\end{figure*}

\begin{figure*}[t]
    \centering
    \includegraphics[width=0.9\textwidth]{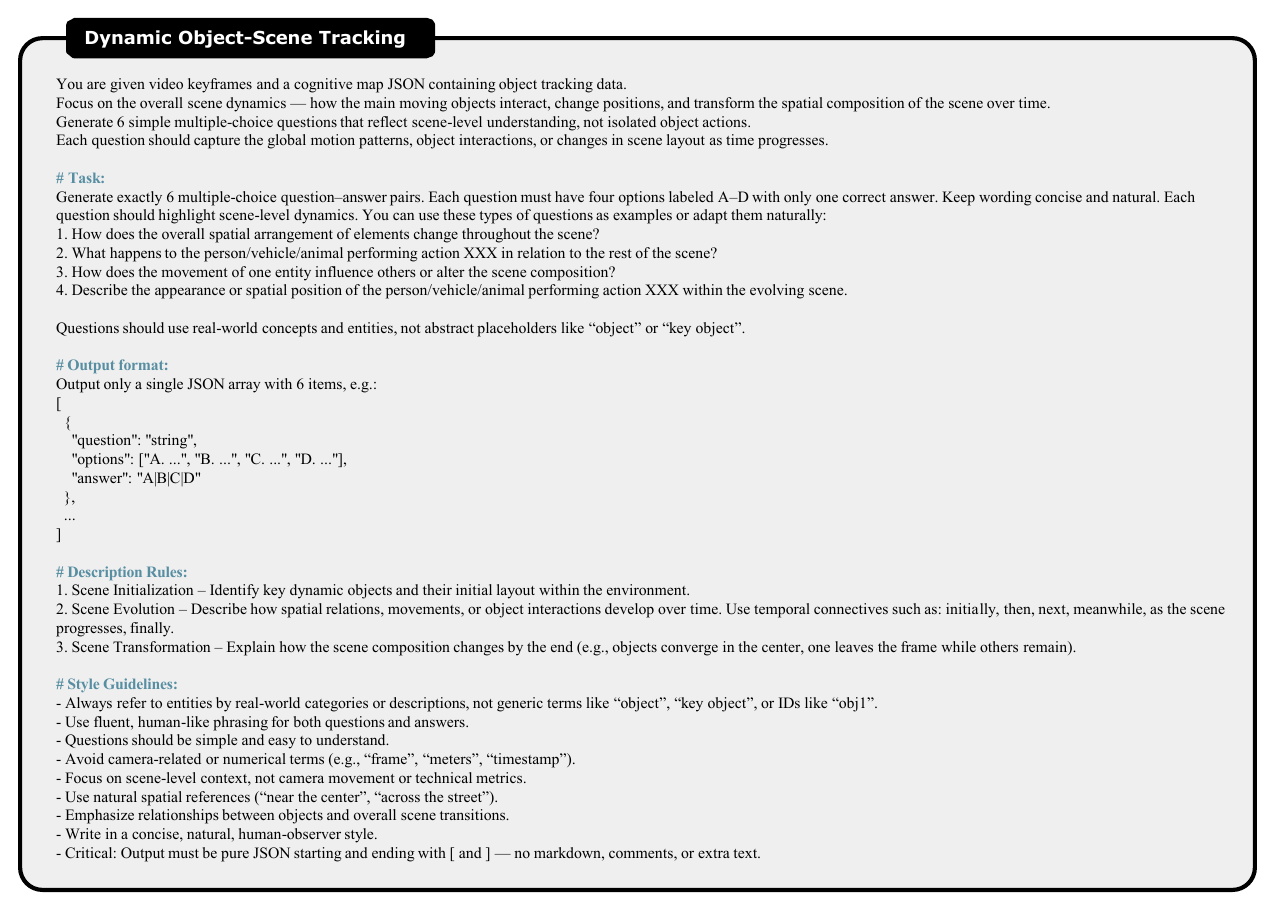}
    \vspace{-2mm}
    \caption{Prompt template for \textit{Dynamic Object-Scene Tracking} in \texttt{Dyn-Bench}. }
    \label{fig:prompt2}
\end{figure*}

\begin{figure*}[t]
    \centering
    \includegraphics[width=0.9\textwidth]{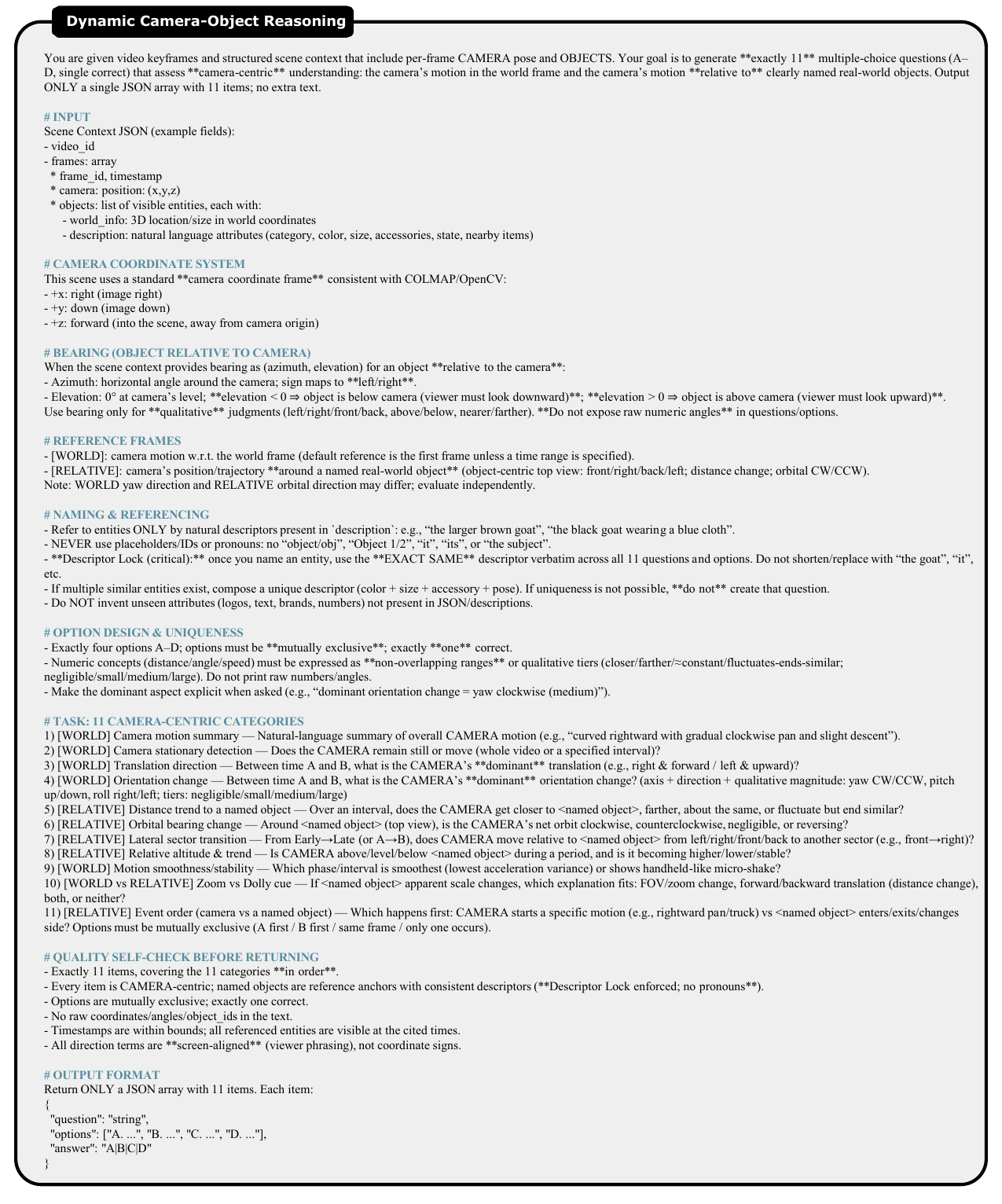}
    \vspace{-2mm}
    \caption{Prompt template for \textit{Dynamic Camera-Object Reasoning} in \texttt{Dyn-Bench}. }
    \label{fig:prompt3}
\end{figure*}

\begin{figure*}[t]
    \centering
    \includegraphics[width=0.9\textwidth]{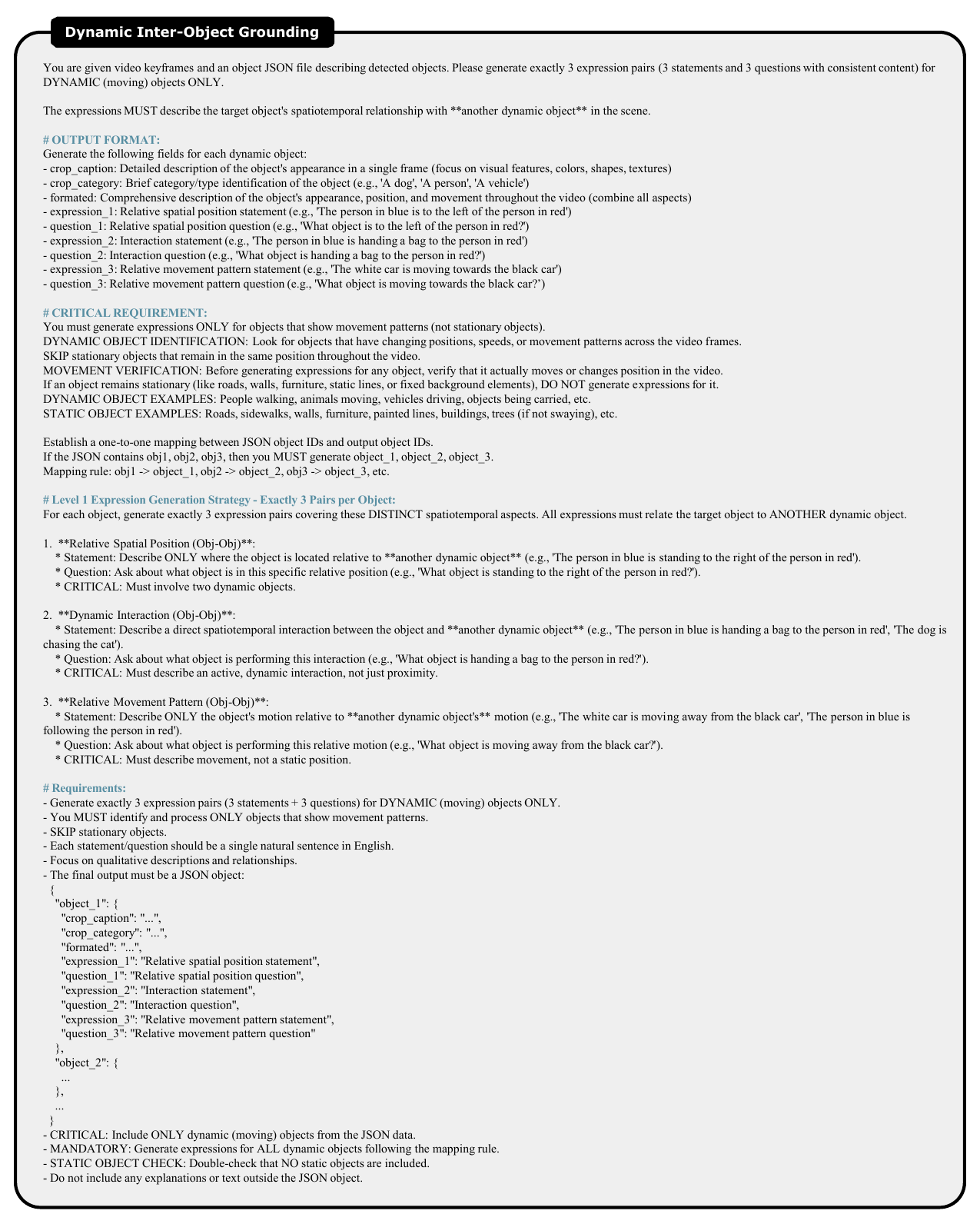}
    \vspace{-2mm}
    \caption{Prompt template for \textit{Dynamic Inter-Object Grounding} in \texttt{Dyn-Bench}. }
    \label{fig:prompt4}
\end{figure*}

\begin{figure*}[t]
    \centering
    \includegraphics[width=0.9\textwidth]{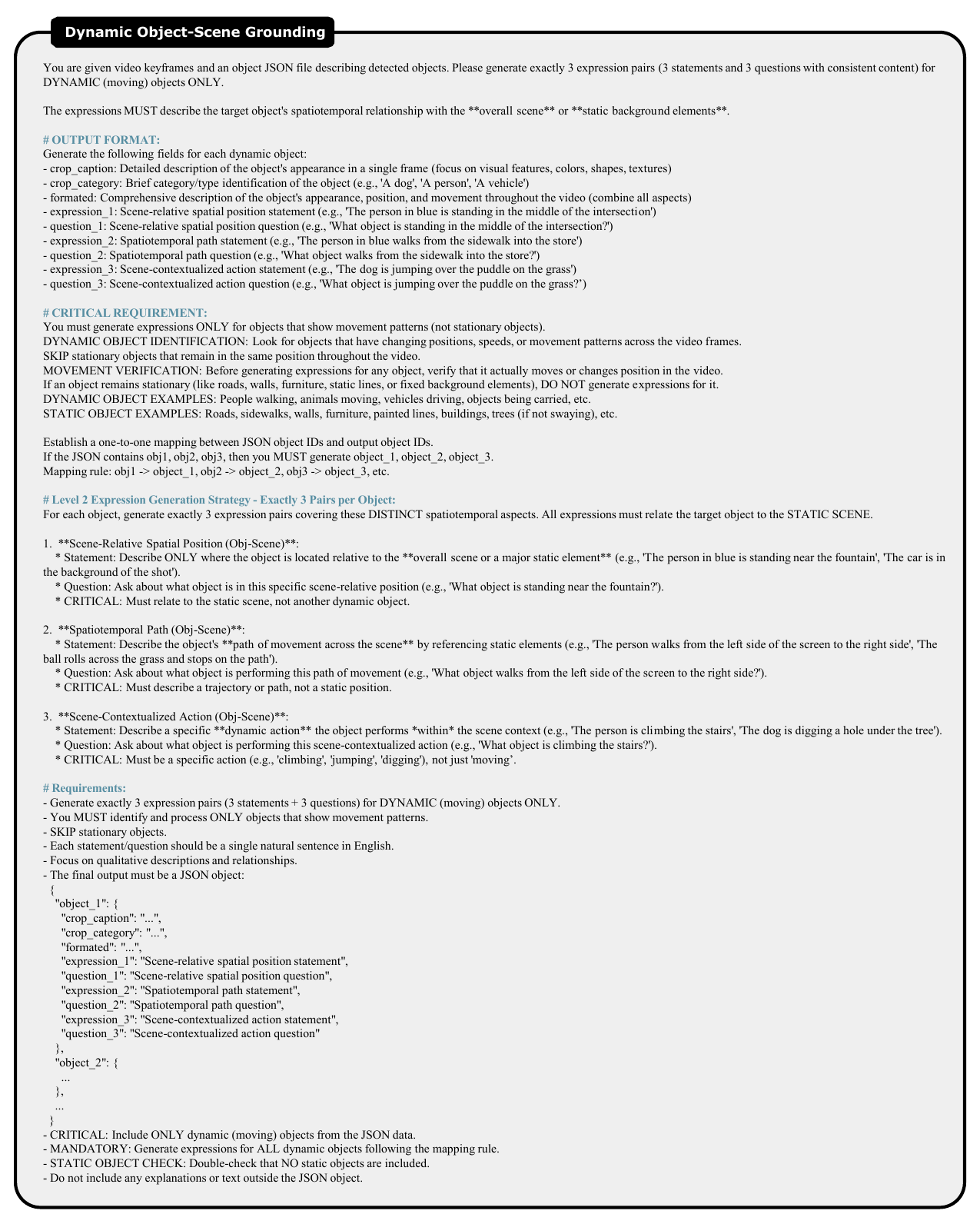}
    \vspace{-2mm}
    \caption{Prompt template for \textit{Dynamic Object-Scene Grounding} in \texttt{Dyn-Bench}. }
    \label{fig:prompt5}
\end{figure*}

\begin{figure*}[t]
    \centering
    \includegraphics[width=0.9\textwidth]{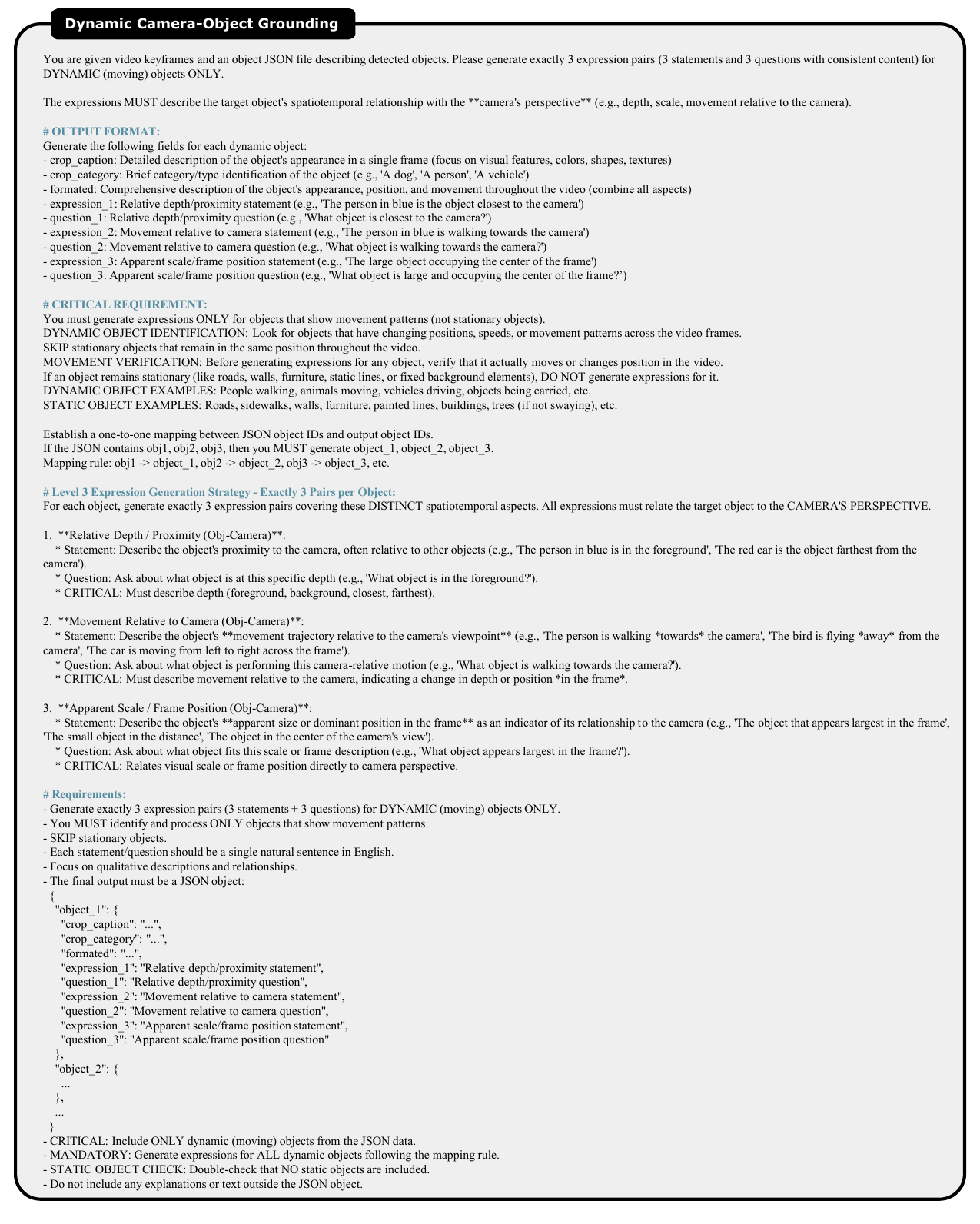}
    \vspace{-2mm}
    \caption{Prompt template for \textit{Dynamic Camera-Object Grounding} in \texttt{Dyn-Bench}. }
    \label{fig:prompt6}
\end{figure*}

We introduce six prompt templates used in the Question–Answer (QA) generation stage for creating both Visual Question Answering (VQA) and grounding pairs. 
Each prompt is provided as input to the \textit{Qwen3-VL-23B} model to elicit diverse, context-aware, and semantically consistent responses. 
The six templates correspond to \textit{Dynamic Inter-Object Perception}, \textit{Dynamic Object-Scene Tracking}, \textit{Dynamic Camera-Object Reasoning}, \textit{Dynamic Inter-Object Grounding}, \textit{Dynamic Object-Scene Grounding}, and \textit{Dynamic Camera-Object Grounding}, as shown in Fig.~\ref{fig:prompt1}–\ref{fig:prompt6}.

\clearpage

\clearpage

\clearpage

\clearpage
\clearpage
\clearpage
\clearpage
\clearpage

% ------------------------

\clearpage

\clearpage
\clearpage
\clearpage
\clearpage
\vspace*{\fill}

\newpage
\vspace*{\fill}
\newpage

% {
%     \small
%     \bibliographystyle{ieeenat_fullname}
%     \bibliography{ref}
% }

\end{document}